\documentclass[10pt,journal,compsoc]{IEEEtran}

\usepackage{times}
\usepackage{epsfig}
\usepackage{graphicx}
\usepackage{amsmath}
\usepackage{amssymb}
\usepackage{overpic}
\usepackage{cleveref}
\usepackage{tabularx}
\usepackage{nicefrac}
\usepackage{booktabs}
\usepackage{bm}
\usepackage{multirow}
\usepackage{mathtools}
\usepackage{amsmath}

\usepackage{amssymb}
\usepackage{amsfonts}
\usepackage{amsopn}
\usepackage{graphicx} 
\usepackage{textcomp}
\usepackage{xfrac}
\usepackage{bbm}
\usepackage{overpic}
\usepackage{subfig}
\usepackage{wrapfig}
\usepackage[ruled,vlined,linesnumbered]{algorithm2e}
\usepackage{multirow}

\usepackage{color}
\definecolor{green}{rgb}{0, 0.5, 0}
\definecolor{orange}{rgb}{0.8, 0.6, 0.2}
\definecolor{red}{rgb}{1.0, 0.0, 0.0}
\definecolor{teal}{rgb}{0.0, 0.4, 0.4}
\definecolor{purple}{rgb}{0.65,0,0.65}
\definecolor{saffron}{rgb}{0.95,0.75,0.2}
\definecolor{turquoise}{rgb}{0.0,0.5,0.5}
\definecolor{brown}{rgb}{0.5, 0.16, 0.16}

\usepackage{overpic}
\usepackage{currfile} 

\newlength\savedwidth

\newcommand{\kx}[1]{{\color{black}#1}}

\newcommand{\revise}[1]{{\color{black}#1}}

\definecolor{lightgray}{rgb}{0.6, 0.6, 0.6}

\usepackage[normalem]{ulem}

\newcommand{\hidecomment}[1]{}

\newcommand{\bX}{\mathbf{X}}

\newcommand{\bc}{\mathbf{c}}

\newcommand{\bF}{\mathbf{F}}
\newcommand{\bV}{\mathbf{V}}

\newcommand{\bW}{\mathbf{W}}
\newcommand{\bz}{\mathbf{z}}
\newcommand{\bh}{\mathbf{h}}
\newcommand{\bQ}{\mathbf{Q}}
\newcommand{\bK}{\mathbf{K}}

\newcommand{\bZ}{\mathbf{Z}}
\newcommand{\bS}{\mathbf{S}}

\usepackage{xspace}

\ifCLASSOPTIONcompsoc
  \usepackage[nocompress]{cite}
\else
  \usepackage{cite}
\fi

\ifCLASSINFOpdf
\else
\fi

\begin{document}
\title{RayMVSNet++: Learning Ray-based 1D Implicit Fields for Accurate Multi-View Stereo}

\author{Yifei Shi,~\IEEEmembership{Member, IEEE,}
        Junhua Xi,
        Dewen Hu,~\IEEEmembership{Senior Member, IEEE,}
        Zhiping Cai,\\
        Kai Xu,~\IEEEmembership{Senior Member, IEEE}
\IEEEcompsocitemizethanks{
\IEEEcompsocthanksitem Yifei Shi and Dewen Hu are with the College of Intelligence Science and Technology, National University of Defense Technology, China.
\IEEEcompsocthanksitem Junhua Xi, Zhiping Cai, and Kai Xu are with the College of Computer Science, National University of Defense Technology, China.
\IEEEcompsocthanksitem Yifei Shi and Junhua Xi contributed equally.
\IEEEcompsocthanksitem Dewen Hu (dwhu@nudt.edu.cn) and Kai Xu (kevin.kai.xu@gmail.com) are the corresponding authors.}}

\markboth{IEEE TRANSACTIONS ON PATTERN ANALYSIS AND MACHINE INTELLIGENCE}%
{Shell \MakeLowercase{\textit{et al.}}: Bare Demo of IEEEtran.cls for Computer Society Journals}

\IEEEtitleabstractindextext{%
\begin{abstract}
Learning-based multi-view stereo (MVS) has by far centered around 3D convolution on cost volumes. Due to the high computation and memory consumption of 3D CNN, the resolution of output depth is often considerably limited. Different from most existing works dedicated to adaptive refinement of cost volumes, we opt to directly optimize the depth value along each camera ray, mimicking the range (depth) finding of a laser scanner. This reduces the MVS problem to ray-based depth optimization which is much more light-weight than full cost volume optimization. In particular, we propose RayMVSNet which learns sequential prediction of a 1D implicit field along each camera ray with the zero-crossing point indicating scene depth. This sequential modeling, conducted based on transformer features, essentially learns the epipolar line search in traditional multi-view stereo.
We devise a multi-task learning for better optimization convergence and depth accuracy.
We found the monotonicity property of the SDFs along each ray greatly benefits the depth estimation.
Our method ranks top on both the DTU and the Tanks \& Temples datasets over all previous learning-based methods, achieving an overall reconstruction score of $0.33$mm on DTU and an F-score of $59.48\%$ on Tanks \& Temples.
It is able to produce high-quality depth estimation and point cloud reconstruction in challenging scenarios such as objects/scenes with non-textured surface, severe occlusion, and highly varying depth range.
Further, we propose RayMVSNet++ to enhance contextual feature aggregation for each ray through designing an attentional gating unit to select semantically relevant neighboring rays within the local frustum around that ray. This improves the performance on datasets with more challenging examples (e.g. low-quality images caused by poor lighting conditions or motion blur).
RayMVSNet++ achieves state-of-the-art performance on the ScanNet dataset.
In particular, it attains an AbsRel of $0.058$m and produces accurate results on the two subsets of textureless regions and large depth variation.
\end{abstract}

}

\maketitle

\IEEEdisplaynontitleabstractindextext

\IEEEpeerreviewmaketitle

\ifCLASSOPTIONcompsoc
\IEEEraisesectionheading{\section{Introduction}\label{sec:intro}}
\else
IEEEhowto:kopka\section{Introduction}
\label{sec:intro}
\fi

\IEEEPARstart{L}{earning}-based multi-view stereo has gained a surge of attention since the seminal work of MVSNet~\cite{yao2018mvsnet}.
The core idea of MVSNet and many followup works is to construct a 3D cost volume in the frustum of the reference view through warping the image features of several source views onto a set of fronto-parallel sweeping planes at hypothesized depths.
3D convolutions are then conducted on the cost volume to extract 3D geometric features and regress the final depth map of the reference view.

Existing methods are often limited to low-resolution cost volume since 3D CNN is both computation and memory consuming.
Several recent works proposed to upsample or refine cost volume aiming at increasing the resolution of output depth maps~\cite{gu2020cascade,yang2020cost,cheng2020deep}. Such refinement, however, still needs to trade off between depth and spatial (image) resolutions. For example, CasMVSNet~\cite{gu2020cascade} opts to narrow down the range of depth hypothesis to allow high-res depth estimation, matching the spatial resolution of input RGB. 3D convolution is then confined within the narrow band, thus degrading the efficacy of 3D feature learning.

\begin{figure}[t!] \centering
	\begin{overpic}[width=1.0\linewidth,tics=10]{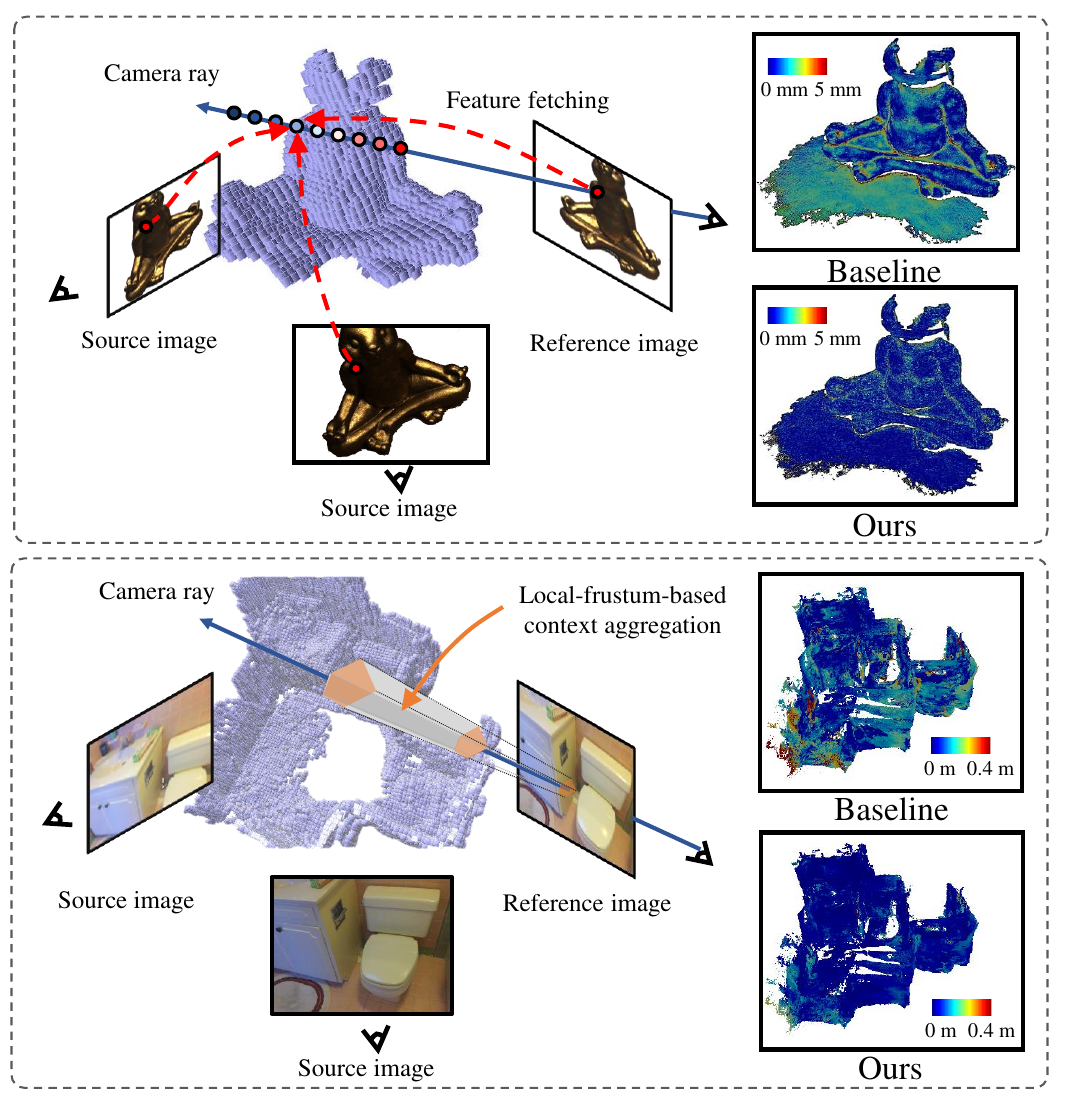}
   \end{overpic}
   \caption{RayMVSNet performs multi-view stereo via predicting 1D implicit fields on a camera ray basis.
    Top: The sequential prediction of 1D field is light-weight and the monotonicity of ray-based distance field around surface-crossing points facilitates robust learning, leading to more accurate depth estimation than the purely cost-volume-based baselines such as MVSNet~\cite{yao2018mvsnet}.
    Bottom: By aggregating more contextual feature with an extra local frustum-based attentional gating unit, RayMVSNet++ is able to achieve more accurate and robust depth predictions in challenging scenarios such as poor lighting conditions or motion blur.
   }
   \label{fig:teaser}
\end{figure}

In fact, depth map is view-dependent although cost volume is not. Since the target is depth map, refining the cost volume seems neither economic nor necessary. There could be a large portion of the cost volume invisible to the view point. We advocate direct optimization of the depth value along each camera ray, mimicking the range (depth) finding of a laser scanner. This allows us to reduce the MVS problem to a ray-based depth optimization one which is, individually, a much more light-weight task than full cost volume optimization. We formulate the ``range finding'' of each camera ray as learning a 1D implicit field along the ray whose zero-crossing point indicates the scene depth along that ray (Figure~\ref{fig:teaser}, top row). To achieve that, we propose RayMVSNet which learns sequential modeling of multi-view features along camera rays based on recurrent neural networks.


Technically, RayMVSNet contains two critical designs to facilitate learning accurate ray-based 1D implicit fields. \emph{Firstly}, the sequential prediction of 1D implicit field along a camera ray is essentially conducting an epipolar line search~\cite{andrew2001multiple} with cross-view feature matching whose optimum corresponds to the point of ray-surface intersection. To learn this line search, we propose \emph{Epipolar Transformer}. Given a camera ray of the reference view, it learns the matching correlation of the pixel-wise 2D features of each source view based on attention mechanism. The transformer features of all views, together with (low-res) cost volume features, are then concatenated and fed into an LSTM~\cite{hochreiter1997long} for implicit field regression. Figure~\ref{fig:epipolar} visualizes how epipolar transformer selects reliable matching features from different views.

\emph{Secondly}, we confine the sequential modeling for each camera ray within a fixed-length range centered around the hypothesized surface-crossing point given by the vanilla MVSNet. This makes the output 1D implicit field along each ray monotonous, which is normalized to $[-1,1]$. Such restriction and normalization lead to significant reduction of learning complexity and improvement of result quality. We devise two learning tasks: 1) sequential prediction of signed distance at a sequence of points sampled in the fixed-length range and 2) regression of the zero-crossing position on the ray. A carefully designed loss function correlates the two tasks. Such multi-task learning approach yields highly accurate estimation of per-ray surface-crossing points.

Learning view-dependent implicit fields has been well-exploited in neural radiance fields (NeRF)~\cite{mildenhall2020nerf} with great success. Recently, NeRF was combined with MVSNet for better generality~\cite{chen2021mvsnerf}. Albeit sharing conceptual similarity, our work is completely different from NeRF. \emph{First}, NeRF (including MVSNeRF~\cite{chen2021mvsnerf}) is designed for novel view synthesis, a different task from MVS. \emph{Second}, the radiance field in NeRF is defined and learned in continuous 3D space and camera rays are used only in the volume rendering stage. In our RayMVSNet, on the other hand, we explicitly learn 1D implicit fields on a camera ray basis.
\emph{Third}, while NeRF is usually trained to fit a given scene, RayMVSNet naturally generalizes to novel scenes.

\kx{
RayMVSNet was published in CVPR 2022~\cite{xi2022raymvsnet} where we demonstrated state-of-the-art performance of RayMVSNet on two public datasets over all learning-based methods.
RayMVSNet achieves an overall reconstruction score of $0.33$mm on DTU and an F-score of $59.48\%$ on Tanks \& Temples.
In particular, RayMVSNet is able to produce high-quality depth estimation and point cloud reconstruction results in challenging scenarios such as objects/scenes with non-textured surface, severe occlusion, and highly varying depth range.
Notably, since all rays share weights for the LSTM and the epipolar transformer, the RayMVSNet model is light weight.
Moreover, the computation for each ray is highly parallelizable.

The ray-based solution, however, has an inherent limitation of insufficient context aggregation; it does not account for the interaction between neighboring rays. This may lead to degraded performance on larger and more complex scenes (such as those from ScanNet~\cite{dai2017scannet}) where context is more essential. In this paper, we propose RayMVSNet++, an augmented version of RayMVSNet, by enhancing the ray-based feature aggregation with \emph{local-frustum-based context aggregation}.
For each ray, we extract its features in the frustum centered around the ray learn.
This amounts to select semantically relevant neighboring rays in the frustum and aggregate the contextual information from those rays.
In particular, an attentional gating unit with the Gumbel-Softmax trick~\cite{jang2016categorical} is designed to make the selection of neighboring rays end-to-end trainable.
This leads to more accurate and robust depth predictions, especially in the challenging scenarios such as poor lighting conditions or motion blur which cannot be well handled by existing methods.

RayMVSNet++ outperforms prior works (including RayMVSNet) on ScanNet, achieving an AbsRel of $0.058$m.
We also demonstrate that RayMVSNet++ is able to produce accurate results on two subsets of ScanNet containing textureless regions and exhibiting large depth variation.

Our work makes the following contributions (those which are newly introduced in RayMVSNet++ are marked with bullet symbol of ``$\ast$''):
\begin{itemize}
  \item A novel formulation of deep MVS as learning ray-based 1D implicit fields.
  \item An epipolar transformer designed to learn cross-view feature correlation with attention mechanism.
  \item A multi-task learning approach to sequential modeling and prediction of 1D implicit fields based on LSTM.
  \item A challenging test set focusing on regions with specular reflection, shadow or occlusion based on the DTU dataset~\cite{aanaes2016large} and associated extensive evaluations.
  \item[$\ast$] A local-frustum-based context aggregation that extends the receptive field of the ray-based model, leading to more accurate and robust predictions.
  \item[$\ast$] New experiments on the ScanNet dataset to comprehensively evaluate the performance in challenging senarios.
\end{itemize}
}

\section{Related work}
\label{sec:related}

\noindent\textbf{Learning-based MVS.}\
Recent advances have made remarkable progress on learning-based MVS. Hartmann et al.~\cite{hartmann2017learned} first propose to learn the multi-patch similarity from two views by a Siamese convolutional network. SurfaceNet~\cite{ji2017surfacenet} and DeepMVS~\cite{huang2018deepmvs} warp the multi-view images into the 3D cost volume and adopt 3D neural networks to estimate the geometry.
LSM~\cite{kar2017learning} introduces differentiable projection operation to enable the end-to-end 3D reconstruction from multi-view images.
MVSNet~\cite{yao2018mvsnet} proposes a differentiable homography and leverages 3D cost volume in a learning pipeline.
MVSNet aggregates contextual information by a 3D convolutional network, especially on regions with complex illumination, specularity, and occlusion.
However, the high computation and memory consumption restrict the output depth resolution, limiting its scalability in large scenes.

To reduce the requirements, many follow-up works have been developed.
R-MVSNet~\cite{yao2019recurrent} proposes to regularize the 2D cost maps along the depth direction so the memory consumption could be greatly reduced.
Point-MVSNet~\cite{chen2019point} first computes the coarse depth with a low-resolution cost volume and then uses a point-based refinement network to generate the high-resolution depth map.
CasMVSNet~\cite{gu2020cascade} adopts a cascade cost volume to gradually narrow the depth range and increase the cost volume resolution.
Similar ideas are later explored to reduce the memory cost of 3D convolutions and/or increase the depth quality, such as coarse-to-fine depth optimization~\cite{yang2020cost,yu2020fast,yan2020dense,xu2020learning,ma2021epp,cheng2020deep,xu2021digging}, attention-based feature aggregation~\cite{luo2020attention,yu2021attention,zhang2021long,wei2021aa}, and patch matching-based method~\cite{luo2019p,wang2021patchmatchnet}.
Unlike these works, RayMVSNet optimizes the depth on each camera viewing ray instead of the 3D volume, which is more light-weight.

Multi-view feature aggregation is one of the most crucial components in learning-based MVS. Previous works adopted various solutions to learn mutual correlations~\cite{zhao2018triangle}, avoiding the influences of incorrect matches caused by occlusion. Popular solutions include the visibility-based aggregation~\cite{zhang2020visibility,chen2020visibility}, the attention-based aggregation~\cite{yi2020pyramid,wei2021aa,yang2022mvs2d}, etc.
RayMVSNet follows the attention-based aggregation route. Nevertheless, it learns feature aggregation at each 3D point, instead of the entire image or volume, thus greatly reducing the memory consumption.

\revise{Our method is also relevant to~\cite{murez2020atlas,sun2021neuralrecon} in terms of reconstructing 3D object by estimating the SDFs. While their methods focus on reconstructing the global TSDF volume of large-scale scenes and generating 3D surfaces with good completeness, our method estimates local SDFs on each camera ray individually resulting in more accurate depth estimation.}

\noindent\textbf{Learning MVS with Transformers.}\
Since the pioneering work of~\cite{vaswani2017attention}, Transformers have significantly advanced the research of natural language processing~\cite{katharopoulos2020transformers,kitaev2020reformer,lee2019set}. More recently, Transformers show great potential in vision tasks, such as image classification~\cite{dosovitskiy2020image,carion2020end}, object detection~\cite{carion2020end}, scene segmentation~\cite{dai2019second}, panoptic segmentation~\cite{li2019attention}, pose estimation~\cite{qin2022geometric}, and visual localization~\cite{sun2021loftr}, thanks to the superb capabilities of modeling long-range dependencies. There are also a bunch of works that utilize Transformers to capture the long-range relations in solving MVS problems. Most of them aggregate context from the extracted 2D image features and solve the cross-view matching problem.
For example, MVSTR~\cite{zhu2021multi}, LANet~\cite{zhang2021long} and TransMVSNet~\cite{ding2022transmvsnet} unitize the attention mechanisms to extract dense features with global contexts. PA-MVSNet~\cite{zhang2021pa} and AACVP-MVSNet~\cite{yu2021attention} introduce self-attention layers for hierarchical features extraction, which is able to capture multi-scale matching clues for the subsequent depth inference task.
AttMVS~\cite{luo2020attention} introduces an attention-enhanced matching confidence volume to improve the matching robustness.
To reduce the searching cost, recent researches~\cite{yang2022mvs2d,he2020epipolar} have been focusing on leveraging the geometric prior of epipolar line by restricting attention associations within the epipolar line, which makes the learning more efficient.
Our method also utilizes the epipolar geometric prior. However, it is different from the previous works as the proposed epipolar transformer essentially learns feature fusion at a 3D point by aggregating multi-view image features, while previous methods learn the matching of 2D pixels from two images. This leads to different network architectures.

\noindent\textbf{Learning Implicit Representation.}\
Many works have attempted learning shape representation based on implicit fields.
Implicit field shows promising results on facilitating a variety number of problems, such as shape reconstruction~\cite{zhang2021learning,zheng2021deep,nelson2020learning,deng2021deformed} and rendering~\cite{mildenhall2020nerf,takikawa2021neural}.
It achieves high quality shape reconstruction by allocating a value to every point in 3D space and extracting the shape surface as an iso-surface.
DeepSDF~\cite{park2019deepsdf} proposes to predict the magnitude of 3D point to indicate the distance to the surface boundary and a sign to determine whether the point is inside or outside of the shape.
IM-Net~\cite{chen2019learning} and Occupancy Network~\cite{mescheder2019occupancy} learn the implicit fields to estimate the point-wise occupancy probability with a binary classifier.
To improve the effectiveness and generalization on complex scenes, latest studies propose to enhance implicit field by introducing extra inputs~\cite{xu2019disn,peng2020convolutional}, adopting advanced learning techniques~\cite{tretschk2020patchnets,duan2020curriculum,niemeyer2020differentiable,sitzmann2020metasdf} and decomposing the scene into local regions~\cite{chabra2020deep,takikawa2021neural,jiang2020local,genova2020local}.
\revise{In particular, PIFu~\cite{saito2019pifu} proposes an implicit representation that locally aligns pixels of 2D images with the global context of their corresponding 3D object.
The method is able to infer both the object surface and texture from single or multiple input images.}

NeRF~\cite{mildenhall2020nerf} represents complex scenes by learning a view-dependent implicit neural radiance field, achieving high-resolution realistic novel view synthesis.
Aside from the reasons mentioned in the introduction, our method is different from NeRF in the following aspects.
First, NeRF learns the radiance field by MLPs. In contrast, our method tackles the problem of cross-view feature correlation with sequential modeling.
Second, our model generalizes to untrained scenes, while NeRF generally does not.
\revise{To increase the NeRF's generality on untrained scenes, a series of methods have been proposed, such as NeuralRay~\cite{liu2022neural}, TransNeRF~\cite{wang2022generalizable}.
In particular, IBRNet~\cite{wang2021ibrnet} learns multi-view image-based rendering with a ray transformer, bringing great cross-scene generality.
Despite the similarity in the concept of inference on the camera ray, our task is different from theirs, resulting in different network designs and training schemes.}

\revise{Since NeRF is designed for view synthesis, it has inferior abilities on approximating the scene's geometry, due to the shape-radiance ambiguities~\cite{zhang2020nerf++}.
Recent works have investigated incorporating the geometric priors or clues, such as the depth prior~\cite{wei2021nerfingmvs} and the TSDF~\cite{yariv2021volume,wang2021neus}, to enhance the scene reconstruction performance while maintaining the quality of view synthesis.
Our method is also different from these methods, as these methods are trained with both appearance and geometry supervision while our method only requires the latter.}


\section{Method}
\label{sec:method}


\begin{figure*}[t] \centering
	\begin{overpic}[width=1.0\linewidth,tics=10]{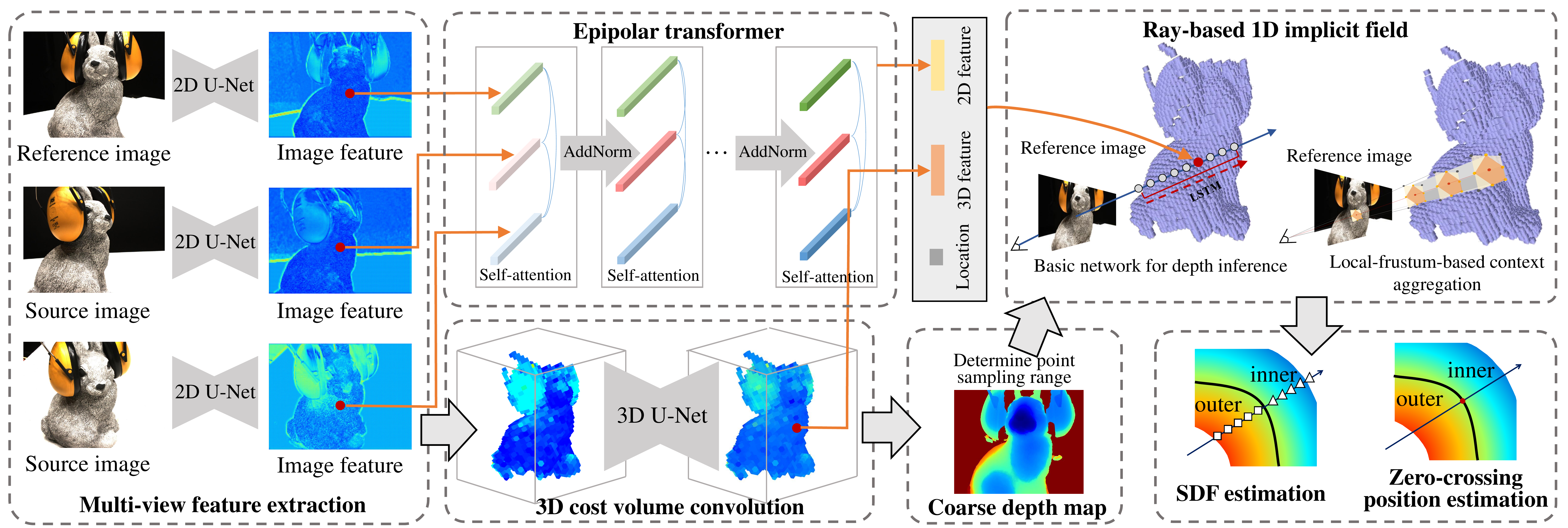}
   \end{overpic}
   \caption{Method overview. Given multiple overlapping RGB images, the multi-view image features are extracted by a 2D U-Net.
   The coarse depth map is then estimated by a coarse 3D cost volume. 2D multi-view image features are then correlated and aggregated by epipolar transformer.
   At last, the ray-based 1D implicit field, which includes a local-frustum-based context aggregation module, is learnt on each camera viewing ray to simultaneously estimate the SDF of the sampled points and the location of the zero-crossing point.
   }
   \label{fig:overview}
\end{figure*} 

\noindent\textbf{Overview.}\
RayMVSNet++ estimates the depth maps from multiple overlapping RGB images. Similar to~\cite{yao2018mvsnet}, at each time, it takes one reference image $I_1$ and $N-1$ source images $\{I_i\}^N_2$ as input, and infers the depth map of the reference image.
RayMVSNet++ starts from building a light-weight 3D cost volume and estimating a coarse depth map (Sec.~\ref{sec:coarse}).
Then, epipolar transformer is proposed to learn the matching correlation of the pixel-wise 2D features of each view using attention mechanism (Sec.~\ref{sec:feature}).
The transformed features are fed into the 1D implicit field, implemented by an LSTM, along each camera viewing ray to estimate the signed distance functions (SDFs) of the hypothesized points as well as the zero-crossing position (Sec.~\ref{sec:ray}).
In particular, a local-frustum-based context aggregation is introduced to aggregate more context from the semantically relevant neighboring rays.
The method overview is illustrated in Figure \ref{fig:overview}.


\subsection{3D Cost Volume and Coarse Depth Prediction}
\label{sec:coarse}
We first feed the multi-view images $\{I_i\}^N_1$ to a 2D U-Net to extract image features $\{\bF^I_i\}^N_1$. The width and height of the image features are the same to those of the input images. Hence, $\{\bF^I_i\}^N_1$ preserve the fine appearance feature of local details, facilitating the high-resolution depth estimation.
By leveraging the 2D multi-view image features and the camera parameters, we build a variance-based 3D cost volume $V$, and extract the 3D volumetric features $\bF^{V}$ via a 3D U-Net~\cite{yao2018mvsnet}.
Since 3D convolution is memory-consuming, the resolution of $V$ in our work is set to be smaller than that in the previous works~\cite{cheng2020deep,yang2020cost,gu2020cascade}.
The coarse depth maps are estimated from the 3D volumetric features, which are then used for determining the modeling range of the ray-based 1D implicit fields.


\begin{figure}[t] \centering
	\begin{overpic}[width=1.0\linewidth,tics=10]{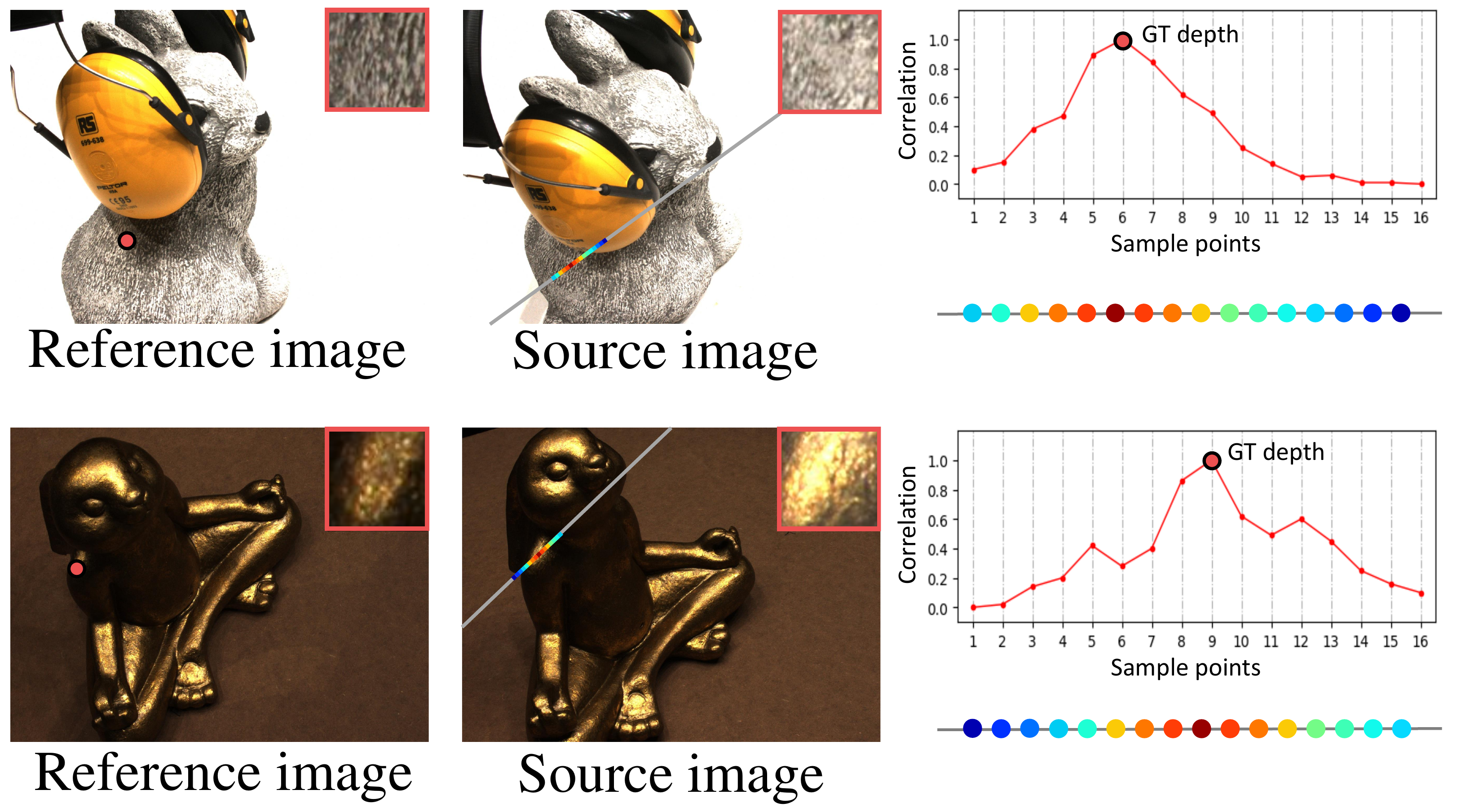}
   \end{overpic}
   \caption{
   Effects of epipolar transformer. Given a point in the reference image, epipolar transformer automatically selects reliable matching feature on the epipolar line of the source image. Note that it finds the matching feature correctly despite the influences of light changing (top row) and specular reflection (bottom row). The visualized point-pair correlations are deduced from the $\text{Softmax}(\bQ\bK^T)$ in Formulation \ref{eq:self_attention}.
   }
   \label{fig:epipolar}
\end{figure} 

\subsection{Epipolar Transformer}
\label{sec:feature}
We cast a set of rays $\mathbf{R}=\{\mathbf{r}_i\}^M_1$ from the camera's viewing direction of the reference image, where $M$ is the number of pixels in the reference image.
Our goal is to estimate the location of the zero-crossing point on each ray, so we can obtain the depth map of the reference view.
Compared to methods that estimate depth on the 3D cost volume, the ray-based method maintains the following advantages.
\emph{First}, since the depth map is view-dependent, ray-based depth optimization is more straightforward and light-weight.
\emph{Second}, all the ray-based 1D implicit fields share an identical spatial property, i.e. the monotonicity of the SDFs along the ray direction. As a result, the learning would be simplified and well regularized, leading to efficient network training and more accurate results.

\begin{figure}[t] \centering
	\begin{overpic}[width=1.0\linewidth,tics=10]{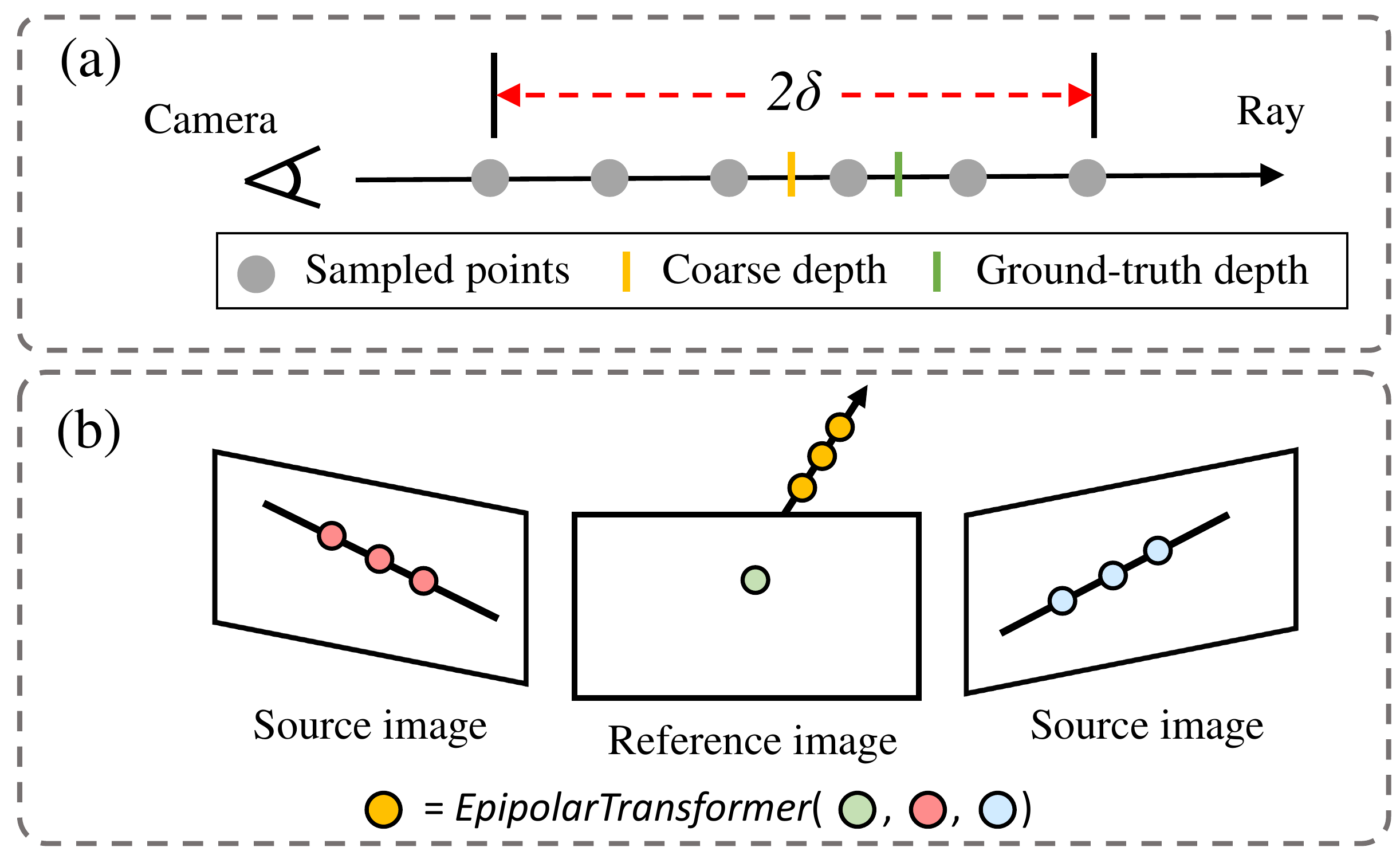}
   \end{overpic}
   \caption{
   (a) The hypothesized points are sampled around the predicted coarse depth to narrow down the search space of the zero-crossing position.
   (b) Epipolar transformer learns the matching correlation of the pixel-wise 2D features and aggregates these features using an attention mechanism.
   }
   \label{fig:method_details}
\end{figure}

\noindent\textbf{Zero-crossing hypothesis sampling.}\
We perform a point sampling to generate the zero-crossing point hypothesis on each ray.
Ideally, one could generate as many points as possible on each ray. However, most of the points are far from the surface, providing less informative information for the depth estimation.
To facilitate efficient training, as shown in Figure~\ref{fig:method_details} (a), we adopt the coarse depth map predicted in sec. \ref{sec:coarse} and uniformly sample $K$ points $P=\{p_k\}^K_1$ on the ray in the range of $\pm \delta$ around the estimated coarse depth.

\noindent\textbf{Attention-aware cross-view feature correlation.}\
The next step is to aggregate feature for the hypothesized points based on the multi-view image features.
A naive way to achieve this is to fetch the features from multi-view images based on the view projection, and take the variance.
However, image feature could be easily influenced by image defects, such as specular reflection and light changing. Naive variance considers all image features equally, which might incur unreliable features and provide incorrect cross-view feature correlation.
To alleviate this problem, we propose \emph{Epipolar Transformer} to learn cross-view feature correlation with attention mechanism (Figure~\ref{fig:method_details} b).

To be specific, the network architecture of epipolar transformer contains four self-attention layers, each followed by two AddNorm layers and one feed-forward layer.
Suppose $\bX=\text{Concat}(\bF^I_{1,p},...,\bF^I_{N,p})$, where $\text{Concat}(\cdot)$ is the concatenation operation, $\{\bF^I_{i,p}\}^N_1$ are the fetched multi-view image features at 3D point $p$. The self-attention layer of epipolar transformer is:
\begin{equation}\label{eq:self_attention}\small
\begin{aligned}
\bS = \text{SelfAttention}(\bQ, \bK, \bV) = \text{Softmax}(\bQ\bK^T)\bV,
\end{aligned}
\end{equation}
where $\bQ=\bX\bW^\bQ$, $\bK=\bX\bW^\bK$, $\bV=\bX\bW^\bV$ are the query vector, the key vector and the value vector respectively. $\bW^\bQ$, $\bW^\bK$, $\bW^\bV$ are the learned weights.
Examples to demonstrate the effects of first self-attention layer in epipolar transformer are visualized in Figure~\ref{fig:epipolar}.
The AddNorm layer of epipolar transformer is:
\begin{equation}\label{eq:addnorm}\small
\begin{aligned}
\bZ = \text{AddNorm}(\bX) = \text{LayerNorm}(\bX+\bS),
\end{aligned}
\end{equation}
where $\text{LayerNorm}(\cdot)$ is the layer normalization operation.
The output of epipolar transformer is the attention-aware denoised multi-view feature $\bF^A_{p}=\{\bF^A_{1,p},...,\bF^A_{N,p}\}$.

To further improve the feature quality, we concatenate the attention-aware feature with the 3D volume feature $\bF^{V}_{p}$ fetched from the 3D cost volume processed in Sec.~\ref{sec:coarse}:
\begin{equation}\label{eq:point_feature2}\small
\bF_p=\text{Concat}(\bF^A_{\mu,p},\bF^A_{\sigma,p},\bF^A_{1,p},\bF^V_{p}).
\end{equation}
where $\bF^A_{\mu,p}$ and $\bF^A_{\sigma,p}$ are the mean and variation of the elements in $\bF^A_{p}$~\cite{yao2018mvsnet,im2019dpsnet}. $\bF^A_{1,p}$ is the attention-aware feature at 3D point $p$ in the reference image.

\subsection{Ray-based 1D Implicit Field}
\label{sec:ray}

\noindent\textbf{LSTM vs. alternative.}\
Given the features of the hypothesized points, the ray-based 1D implicit fields are learned with an LSTM~\cite{hochreiter1997long}.
Crucially, we leverage two attributes of LSTM. \emph{First}, the mechanism of sequential processing inherently facilitates the learning of the SDF monotonicity along the ray direction. \emph{Second}, the property of time invariance increases the network robustness by allowing the zero-crossing position to appear at any place (time-step) on the ray.
An alternative to performing sequential inference is to use transformer~\cite{vaswani2017attention}. However, we experimentally found that replacing LSTM with transformer would not make the performance improve (see Table~\ref{tab:ablation}).
The reason might be that transformer, which is designed for modeling non-local relations, does not explicitly encode relative or absolute position information~\cite{shaw2018self}, making it less suitable to our zero-crossing position searching problem.

\begin{figure}[t] \centering
	\begin{overpic}[width=1.0\linewidth,tics=10]{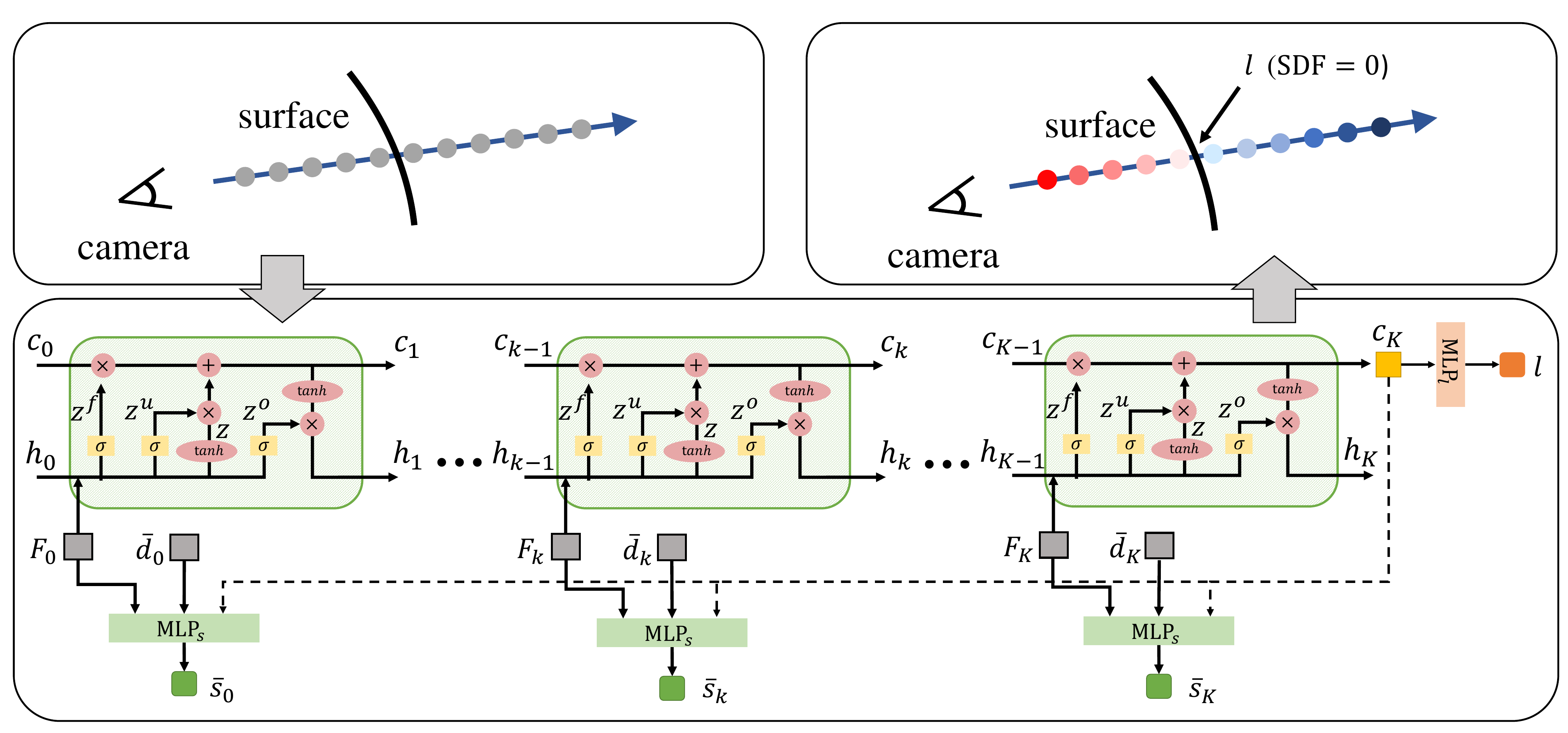}
   \end{overpic}
   \caption{Basic network architecture of the ray-based 1D implicit field. The hypothesized points are fed into an LSTM sequentially, to estimate the position of the zero-crossing point as well as the SDFs.}
   \label{fig:lstm}
\end{figure}

\noindent\textbf{Basic network architecture.}\
The network architecture of the 1D implicit field is shown in Figure~\ref{fig:lstm}.
The LSTM first aggregates the hypothesized points sequentially, and generates the ray feature $\bc_K$.
Specifically, the formulations of an LSTM unit at time-step $k$ are:
\begin{equation}\label{eq:lstm}\small
\begin{aligned}
&\bz=\text{tanh}(\bW[\bF_k,\bh_{k-1}]+b),\\
&\bz^f=\sigma(\bW^f[\bF_k,\bh_{k-1}]+b^f),\\
&\bz^u=\sigma(\bW^u[\bF_k,\bh_{k-1}]+b^u),\\
&\bz^o=\sigma(\bW^o[\bF_k,\bh_{k-1}]+b^o),\\
&\bc_k=\bz^f\circ \bc_{k-1}+\bz^u\circ \bz,\\
&\bh_k=\bz^o\circ \text{tanh}(\bc_k),\\
\end{aligned}
\end{equation}
where $\bF_k$ is the feature of point $p_k$, $\bh_k$ and $\bh_{k-1}$ are the hidden state of point $p_k$ and $p_{k-1}$ respectively, $\bz$ is the cell input activation vector, $\bz^f$ is the activation vector of the forget gate, $\bz^u$ is the activation vector of the update gate, $\bz^o$ is the activation vector of the output gate, $\bc_k$ is the cell state vector, $\bW$, $\bW^f$, $\bW^u$, $\bW^o$ are the weight matrices, $b$, $b^f$, $b^u$, $b^o$ are the weight vectors, $\circ$ is the element-wise multiplication, $\sigma(\cdot)$ is the sigmoid function. The LSTM is initialized with $\bc_0=0$ and $\bh_0=0$.


For each hypothesized point $p_k$, we use the ray feature $\bc_K$, the point-wise feature $\bF_k$ and its depth value $d_k$ (indicating the location on the ray) to estimate its SDF $s_k$ using an MLP.
Instead of using the true depth value $d_k$ and estimating the true SDF $s_k$, we use the normalized depth value $\overline{d}_k=k/K$ and the normalized SDF $\overline{s}_k=s_k/s_{max}\in[-1,1]$, where $s_{max}$ is the maximal absolute SDF value on the ray.
Such normalization leads to a significant reduction of learning complexity and improvement of the result quality.
The formulation of the SDF prediction is:

\begin{equation}\label{eq:sdf}\small
\overline{s}_k=\text{MLP}_s([\bc_K, \bF_k, \overline{d}_k]).
\end{equation}

The above network predicts the SDFs of the hypothesized points on the ray. However, post-processing, e.g. ray casting, is still needed to find the zero-cross position. 
We extend our method to estimate the zero-cross position explicitly with another MLP.
Taking the ray feature $\bc_K$ as input, the MLP predicts the zero-crossing location $l$ on the ray in the normalized 1D coordinate:

\begin{equation}\label{eq:ratio}\small
l=\text{MLP}_l(\bc_K).
\end{equation}



\begin{figure*}[t] \centering
	\begin{overpic}[width=1.0\linewidth,tics=10]{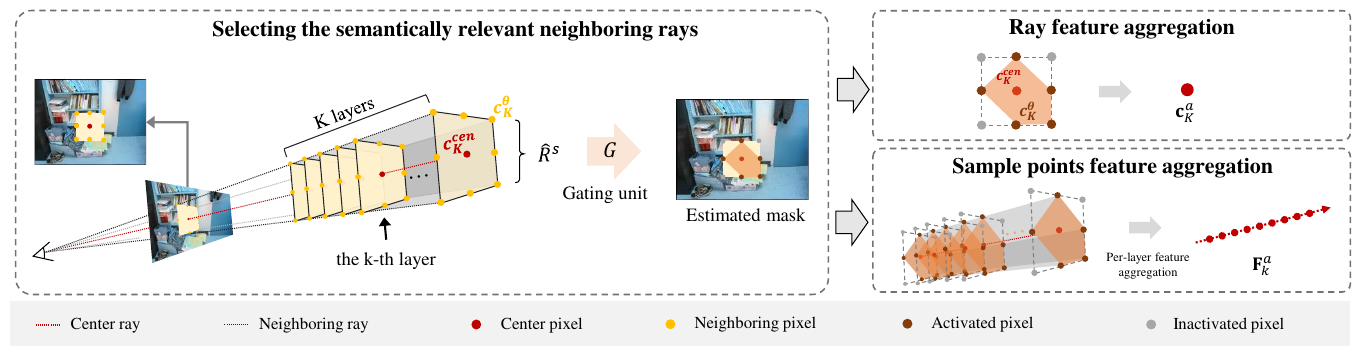}
   \end{overpic}
   \caption{
   Attentional gating unit for the local-frustum-based context aggregation.
   For each camera rays of the reference image, the method estimates a mask that denotes the semantically relevant neighbouring rays. Based on the mask, ray feature and sample points feature, with more contextual information, are aggregated respectively.
   }
   \label{fig:gating1}
\end{figure*} 

\begin{figure}[t] \centering
	\begin{overpic}[width=1.0\linewidth,tics=10]{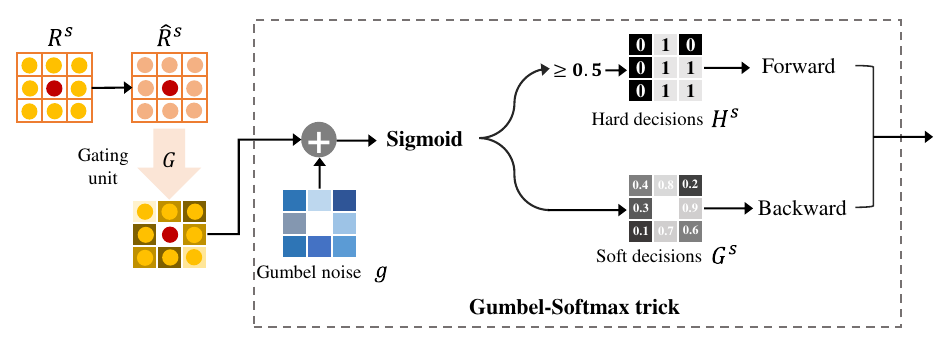}
   \end{overpic}
   \caption{
   Training to estimate the mask with the Gumbel-Softmax trick. The gating unit $G$ (with adding the Gumbel noise) generates the soft gating decisions $G^s$ . The soft decisions are converted into hard decisions during the forward pass. The soft decisions are retained for the backward pass, making the gating unit differentiable.}
   \label{fig:gating2}
\end{figure} 

\noindent\textbf{Local-frustum-based context aggregation.}\
The basic network architecture described above focuses on the inference along each ray direction. This method would work in scenarios where the images are clearly captured under satisfactory conditions, e.g. in good lighting and without motion blur.
This is because in such scenarios the features aggregated along the ray direction are able to provide sufficient information to infer the underlying geometry.
Nevertheless, there is a flurry of data~\cite{silberman2012indoor,dai2017scannet} that does not meet these requirements, making the depth estimation either inaccurate or infeasible.
As such, specific mortifications should be taken to allow the method to tolerate those disadvantages.

We tackle this problem by proposing a simple yet effective method:
consider the interaction between neighboring rays and aggregate more contextual feature to boost ray-based inference.
To achieve this, based on the basic network architecture above, we introduce a local-frustum-based context aggregation module that adaptively aggregates contextual features from neighbouring rays.
By involving more context, the ray feature $\bc_K$ and 3D point feature $\bF_k$ in formulation~(\ref{eq:sdf}) and formulation~(\ref{eq:ratio}) are expected to be more informative and thus result in more accurate depth estimation.

To achieve this, we first extract the features of each ray individually by the above LSTM.
By projecting the ray features to the corresponding pixels in the reference image, we generate a feature map whose width and height are the same as those of the reference image.
For any pixel in the feature map, we set its receptive field as a square with width $t$, and the center is the pixel.
Suppose $\bc_K^\text{cen}\in \mathbb{R}^{\kappa}$ is the extracted feature of the center pixel. $\kappa$ is the feature-length.
$\{\bc_K^\theta\},\theta\in (1,\Theta)$ are the extracted feature of the neighbouring pixels, where $\Theta=(t+1)^2-1$ is the number of neighbouring pixels.

A naive solution to aggregate context in the square is using average-pooling or max-pooling.
However, as not all neighboring rays are equally important to the central ray, the naive pooling would involve irrelevant features and therefore debilitate the network training.
To address this problem, we introduce an \emph{attentional gating unit} (see Figure~\ref{fig:gating1}) that dynamically selects semantically relevant neighboring rays within the local frustum and adaptively aggregates their features, conditioned on the extracted ray-wise features.

We first consider the variance of $\bc_K^\text{cen}$ and $\{\bc_K^\theta\},\theta\in (1,\Theta)$, and generate a tensor $\hat{R}^s = \{\bc_K^\theta-\bc_K^\text{cen}\},\theta\in (1,\Theta)$.
$\hat{R}^s\in \mathbb{R}^{\kappa\times\Theta}$ is taken as the input to the gating unit $G$, estimating the soft gating decisions $G^s\in \mathbb{R}^{\Theta}$ that indicates how relevant are each ray to the central ray:
\begin{equation}\label{eq:gating1}\small
G^s=\sigma(G(\hat{R}^s)+g),
\end{equation}
where $g$ is the Gumbel noise,
the gating unit $G$ is implemented as a 1D MLP in our method.
Note that although we do not use any semantic supervision directly, we found that most of the selected pixels have the same semantic labels as the center pixel (see Figure~\ref{fig:frustum_vis}). The reason is that both $\bc_K^\text{cen}$ and $\{\bc_K^\theta\}$ are high-level features which already contain semantic information.

Then, a Gumbel-Softmax module $H$~\cite{jang2016categorical} turns soft decisions $G^s$ into hard decisions $H^s\in \{0,1\}^{\Theta}$ by replacing the softmax with an argmax during the forward pass and retaining the softmax during the backward pass~\cite{kong2019pixel,verelst2020dynamic}:
\begin{equation}\label{eq:gating2}\small
H^s=H(G^s).
\end{equation}
The hard decision $H^s$ is a binary mask that indicates which ray is semantically relevant to the center ray.
The Gumbel-Softmax module provides a mechanism that outputs a binary mask in the forward pass and also allows the gradient to be back-propagated in the backward pass.
As is shown in Figure~\ref{fig:gating2}, the gating attentional unit is end-to-end trainable.


\begin{figure}[t] \centering
	\begin{overpic}[width=1.0\linewidth,tics=10]{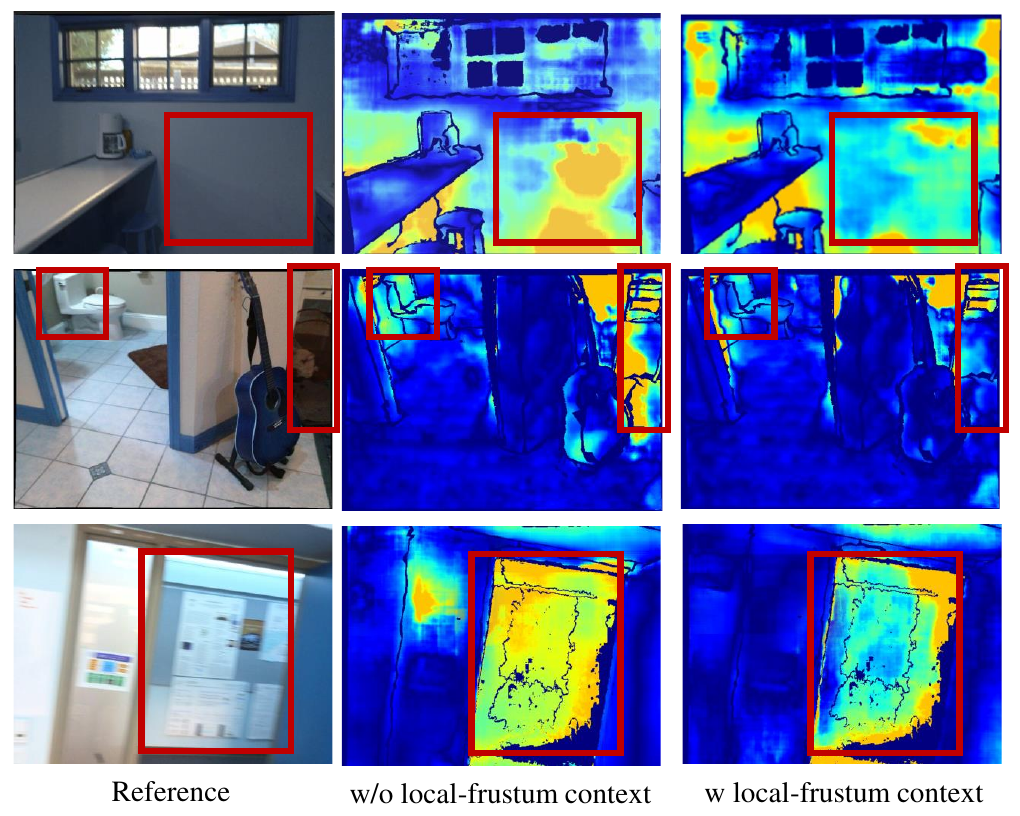}
   \end{overpic}
   \caption{Visual comparison of depth estimation with and without the local-frustum-based context aggregation. Please pay attention to the results of the challenging areas highlighted in the figure.}
   \label{fig:context}
\end{figure} 

Having determined the semantically relevant neighboring rays, we then aggregate context on the activated positions in the mask.
We consider contextual feature aggregation from two aspects.

For \emph{ray feature aggregation}, we take the features from the activated positions, take the average, and add it to the initial central ray feature:
\begin{equation}\label{eq:aggregation12}\small
\bc_K^a = \frac{\sum_{\theta=1}^\Theta (R^s\circ H^s)}{\left \| H^s\right\|_0}\oplus \bc_K,
\end{equation}
where $\bc_K^a\in \mathbb{R}^\kappa$ is the aggregated feature of the central ray, $R^s = \{\bc_K^\theta\},\theta\in (1,\Theta)$ are the features of the neighbouring rays, $\circ$ is the element-wise multiplication, $\oplus$ is the element-wise addition,
$\left \| H^s\right\|_0$ is the number of activated pixel in $H^s$.

For \emph{sample points feature aggregation}, we adopt the same mask and aggregate feature at the $k$-th layer of the frustum:
\begin{equation}\label{eq:aggregation}\small
\bF_k^a = \frac{\sum_{\theta=1}^\Theta (P^s_k\circ H^s)}{\left \| H^s\right\|_0}\oplus \bF_k,
\end{equation}
where $\bF_k^a\in \mathbb{R}^{\kappa}$ is the aggregated feature of the $k$-th sampled point in the central ray,
$P^s_k=\{F_k^\theta\},\theta\in(1,\Theta)$ is the feature map of the neighbouring rays at the $k$-th layer of the frustum.

Last, we replace the $\bc_K$ and $\bF_k$ by $\bc_K^a$ and $\bF_k^a$, respectively, in formulation~(\ref{eq:sdf}) and formulation~(\ref{eq:ratio}).
Therefore, the SDF prediction and zero-crossing location are turned to be:
\begin{equation}\label{eq:sdf_ratio}\small
\begin{aligned}
\overline{s}_k=&\text{MLP}_s([\bc_K^a, \bF_k^a, \overline{d}_k]),\\
l=&\text{MLP}_l(\bc_K^a).
\end{aligned}
\end{equation}

This local-frustum-based context aggregation improves the performance on datasets where challenging regions and low-quality images exist.
The low-quality images are typically captured due to motion blur or bad lighting conditions, which cannot be well handled by existing methods.
We found that the attentional gating unit, without any semantic supervision, tends to select the pixels that belong to the same object as the central pixel.
That is why we claim that the proposed method is able to select semantically relevant neighboring rays for context aggregation.
Please see a visual illustration of its effects in Figure~\ref{fig:context}.





\noindent\textbf{Loss functions.}\
We adopt a multi-task learning strategy to optimize the network. The two tasks, i.e. SDF estimation and zero-crossing position estimation, are inherently relevant and could reinforce each other by optimizing the following loss:

\begin{equation}\label{eq:loss_all}\small
\begin{aligned}
\mathcal{L} &= w_s\mathcal{L}_s + w_l\mathcal{L}_l + w_{sl}\mathcal{L}_{sl},\\
\end{aligned}
\end{equation}
where $\mathcal{L}_s$ and $\mathcal{L}_l$ are the loss of the SDF estimation and the zero-crossing location estimation, respectively:

\begin{equation}\label{eq:loss_s_l}\small
\begin{aligned}
\mathcal{L}_s &= \sum_{k=1}^K L_1(s_k,\hat{s_k}),\\
\mathcal{L}_l &= L_1(l,\hat{l}),\\
\end{aligned}
\end{equation}
where $\hat{s_k}$ and $\hat{l}$ are the ground-truth, $L_1(\cdot)$ denotes the L1 loss function. $\mathcal{L}_{sl}$ is a relational loss that penalizes the inconsistency between the predicted SDFs and the predicted zero-crossing position:
\begin{equation}\label{eq:loss_sl}\small
\begin{aligned}
\mathcal{L}_{sl} &=
\begin{cases}
1, & s_l^a \times s_l^b \textgreater 0\\
0, &  s_l^a \times s_l^b \leq 0,
\end{cases}
\end{aligned}
\end{equation}
where $s_l^a$ and $s_l^b$ are the predicted SDF of the closest two sampled points around the predicted zero-crossing position on the ray.
$w_s$, $w_l$, $w_{sl}$ are the pre-defined weights.


\subsection{Implementations}
We provide implementation details of the training and inference.
The input image size are $640\times 512$, $1600\times 1200$, and $640\times 480$ for the DTU, the Tanks $\&$ Temples, and the ScanNet datasets, respectively.
The 2D U-Net consists of $6$ convolutional layers and $6$ deconvolutional layers, each followed by a batch normalization layer and a ReLU layer, except for the last ones.
The 3D cost volume is fed into a 3D U-Net which consists of three 3D convolutional layers and three 3D deconvolutional layers.
On each ray, the number of hypothesized points $K$ is $16$.
The feature fetching from images and volume are achieved by using bilinear interpolation and trilinear interpolation, respectively.
The hidden dimension of $\bz, \bz^f, \bz^u, \bz^o, \bc_k, \bh_k$ are $50$.
$\text{MLP}_l$ and $\text{MLP}_s$ both contain $4$ fully-convolutional layers. The weights $w_s$, $w_l$, $w_{sl}$ of multi-task learning loss function are $0.1$, $0.8$, $0.1$, respectively. Epipolar transformer and the LSTM are jointly trained. We use Adam optimizer with initial learning rate $0.0005$ which is decreased by $0.9$ for every $2$ epochs.
The training takes $48$ hours. The inference time is about $2$ seconds.
We filter and fuse the depth maps to produce 3D point cloud like previous work~\cite{yao2018mvsnet}.
The receptive field $t$ of the local-frustum-based context aggregation is $9$.
During the training of the attentional gating unit, we use a similar strategy to~\cite{kong2019pixel} that penalizes the trivial solution, e.g. simply using all the neighboring pixels.
We found this strategy would greatly facilitate the training.
The RayMVSNet++ is trained and tested on an NVIDIA Tesla V100-SXM2.

\section{Results and Evaluation}
\label{sec:result}

\subsection{Datasets and evaluation Metrics}
We performed a series of experiments on multiple datasets to evaluate how well our method performs on different scenarios.
The experimental datasets are:



\begin{itemize}
\item \emph{DTU}~\cite{aanaes2016large}: The DTU dataset contains $79$ training scans and $22$ testing scans, all captured under changing lighting conditions.
    Since DTU did not provide SDF annotations, we densely generate the point-wise SDFs from the reconstructed surfaces~\cite{yao2018mvsnet,park2019deepsdf}. Besides, three challenging test subsets focusing on regions with \emph{Specular reflection}, \emph{Shadow} and \emph{Occlusion} are created from the DTU test set. These regions are manually annotated and are designed for evaluating the method's performance in challenging cases. Please refer to the supplemental material for the subsets details.
\item \emph{Tanks \& Temples}~\cite{knapitsch2017tanks}: To evaluate the generality, we test our method on the Tanks \& Temples dataset which contains large-scale complex scenes, using the trained model on \emph{DTU} without any fine-tuning.
\item \emph{ScanNet}~\cite{dai2017scannet}: The ScanNet dataset is originally collected for the purpose of RGB-D reconstruction and scene understanding.
    Since the images are captured in various indoor scenes under ordinary conditions, we utilize the ScanNet dataset for examining the methods' ability on data with low-quality images. Specifically, we collect $31,051$ image triples for training and $1,467$ image triples for testing. The test set could be divided into two subsets: \emph{Textureless} and \emph{Large depth variation}. The point-wise SDFs are generated from the reconstructed surfaces~\cite{yao2018mvsnet,park2019deepsdf}.

\end{itemize}
The statistics of the experimental datasets are reported in Table~\ref{tab:statistics}.

\begin{table}
\footnotesize\centering
\caption{
Statistics of the experimental datasets.
}
\label{tab:statistics}
\setlength{\tabcolsep}{3pt}
\begin{tabular}{ccccc}
\toprule
Dataset & Subset&	\#View&	\#Object&	\#Scene \\
\midrule
\multirow{5}{*}{DTU~\cite{aanaes2016large}}& Train&	3871&	79&	-     \\
& Test&	1078&	22&	-   \\
& Specular reflection&	233&	5&	-   \\
& Shadow&	294&	6&	-   \\
& Occlusion&	254&	5&	-   \\
\midrule
\multirow{4}{*}{ScanNet~\cite{dai2017scannet}}& Train&	31051&	-&	851     \\
& Test&	967&	-&	401   \\
& Textureless&	329&	-&	117   \\
& Large depth variation&	171&	-&	80   \\
\midrule
Tanks and Temples~\cite{knapitsch2017tanks}& Test&	2100&	8&	- \\
\bottomrule
\end{tabular}
\end{table}

\begin{table}
\footnotesize\centering
\caption{
Quantitative results on the \emph{DTU} dataset. We compare all methods using the distance metric~\cite{aanaes2016large}. The numbers are reported in $mm$ (lower is better).
}
\label{tab:dtu}
\begin{tabular}{cccc}
\toprule
Method & Accuracy & Completeness & Overall \\ \midrule
Gipuma~\cite{galliani2015massively} & 0.283 & 0.873 & 0.578\\
MVSNet~\cite{yao2018mvsnet} & 0.396 & 0.527 & 0.462\\
R-MVSNet~\cite{yao2019recurrent} & 0.383 & 0.452 & 0.417\\
CIDER~\cite{xu2020learning} & 0.417 & 0.437 & 0.427\\
P-MVSNet~\cite{luo2019p} & 0.406 & 0.434 & 0.420\\
Point-MVSNet~\cite{chen2019point} & 0.342 & 0.411 & 0.376\\
Fast-MVSNet~\cite{yu2020fast} & 0.336 & 0.403 & 0.370\\
Att-MVSNet~\cite{luo2020attention} & 0.383 & 0.329 & 0.356\\
CasMVSNet~\cite{gu2020cascade} & 0.325 & 0.385 & 0.355\\
CVP-MVSNet~\cite{yang2020cost} & 0.296 & 0.406 & 0.351\\
PatchmatchNet~\cite{wang2021patchmatchnet} & 0.427 & 0.277 & 0.352\\
UCS-Net~\cite{cheng2020deep} & 0.338 & 0.349 & 0.344\\
AACVP-MVSNet~\cite{yu2021attention} & 0.357 & 0.326 & 0.341\\
U-MVS~\cite{xu2021digging} & 0.354 & 0.353 & 0.354\\ \midrule
RayMVSNet & 0.341 & 0.319 & 0.330\\
RayMVSNet++ & 0.344 & 0.312 & \textbf{0.328}\\ \bottomrule
\end{tabular}
\end{table}
\vspace{7pt}

We use the following metrics to evaluate the performances on different datasets, respectively:
\begin{itemize}
\item \emph{Accuracy \& Completeness}~\cite{seitz2006comparison}: the metric that evaluates the accuracy of the reconstructed points (i.e. how close the reconstructed points are to the ground-truth surface) and the completeness of the reconstructed points (i.e. how much of the ground-truth surface is modeled by the reconstructed points). Besides, an overall score is computed as the mean of the accuracy and completeness to indicate the performance considering both the two factors.  We use this metric to evaluate the performance on \emph{DTU}.
\item \emph{F-score }~\cite{knapitsch2017tanks}: the metric that evaluates the precision and recall of the reconstructed points with a specific distance threshold. We use this metric to evaluate the performance on \emph{Tanks \& Temples}. The distance thresholds are different for the tested scenes according to~\cite{knapitsch2017tanks}. F-score is different to the overall score in Accuracy \& Completeness, as it uses the harmonic mean, instead of the arithmetic mean, of precision and recall, resulting in a more balanced metric for measuring the two factors at the same time.
\item \emph{Depth accuracy}~\cite{lee2019big}: we use several metrics for evaluating the estimated depth map comprehensively. The metrics include: AbsRel, SqRel, Log10, RMSE, RMSELog, $\delta<1.25$, $\delta<1.25^2$ , $\delta<1.25^3$, and Percentage$@$x. Table~\ref{tab:depth_metric} reports the details. We use this metric to evaluate the performance on \emph{DTU} and \emph{ScanNet}.
\end{itemize}

\subsection{Performance on DTU}




\begin{figure}[t] \centering
	\begin{overpic}[width=1.0\linewidth,tics=10]{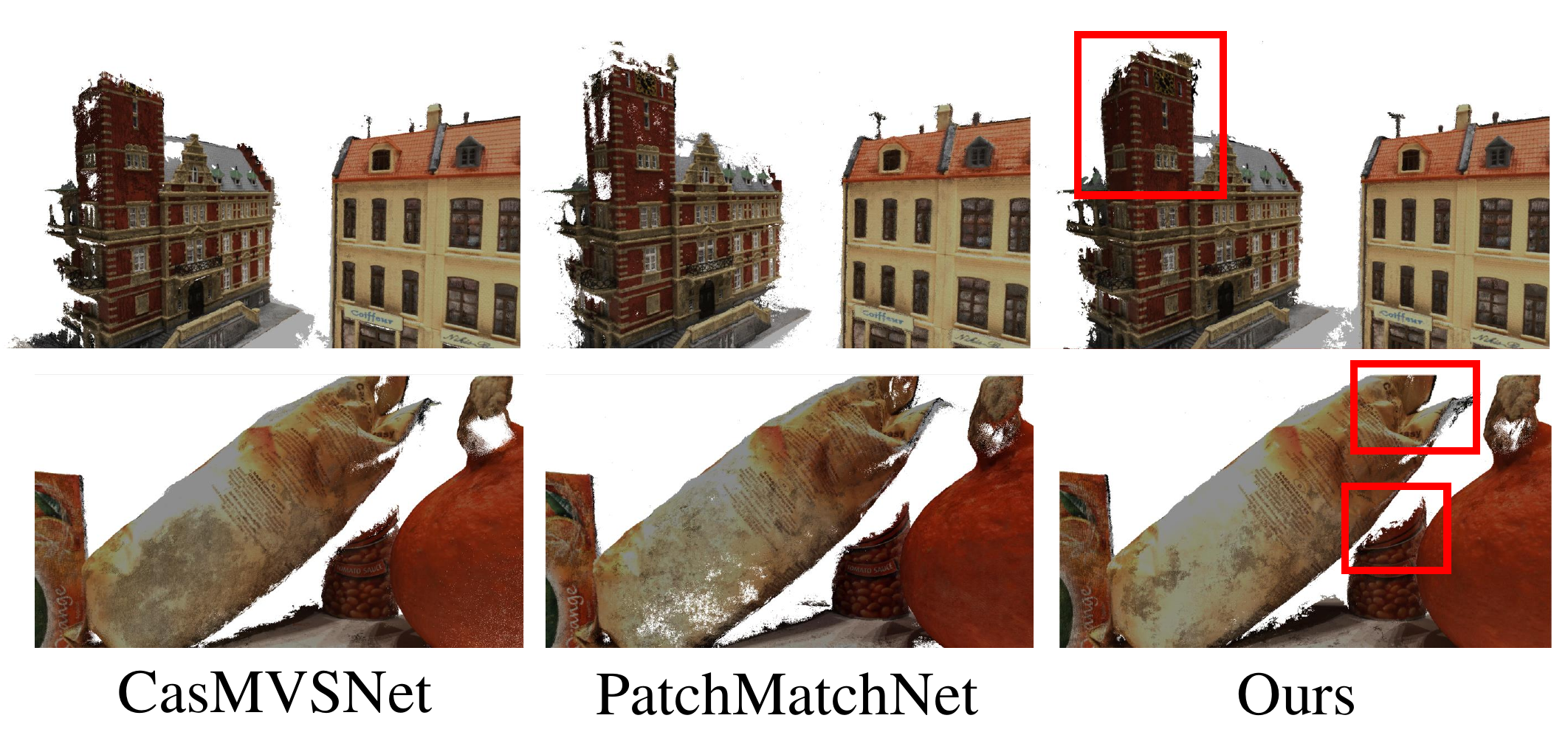}
   \end{overpic}
   \caption{Visual comparison of the reconstructed point cloud by RayMVSNet and the baselines.
   Please pay attention to the results of the challenging areas highlighted in the figure.}
   \label{fig:visual_comp}
\end{figure} 

\begin{figure}[t] \centering
	\begin{overpic}[width=1.0\linewidth,tics=10]{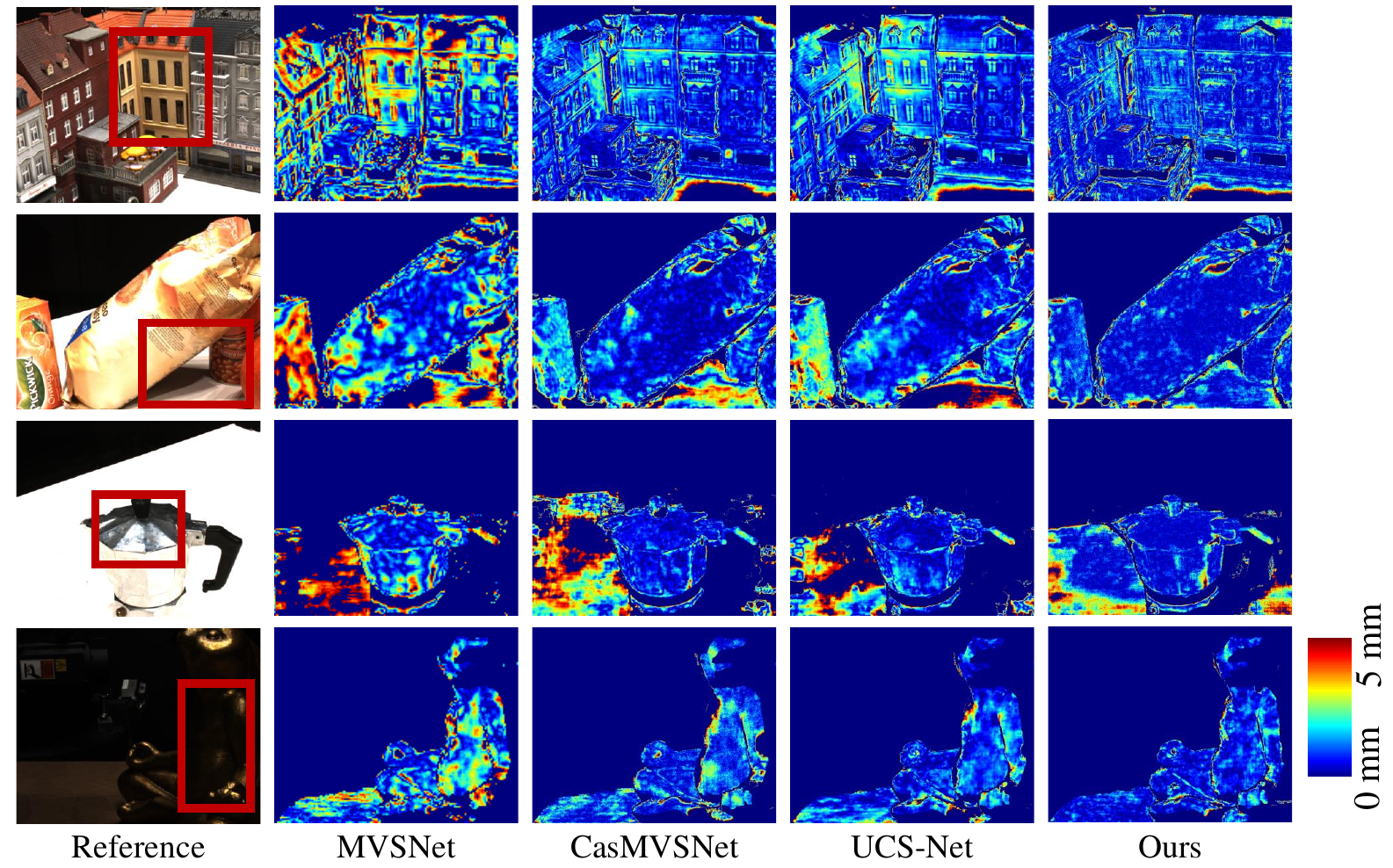}
   \end{overpic}
   \caption{
   Visual comparison of the estimated depth map by RayMVSNet and the baselines. Please pay attention to the results of the challenging areas highlighted in the figure.}
   \label{fig:depth}
\end{figure}


\begin{figure}[t] \centering
	\begin{overpic}[width=1.0\linewidth,tics=10]{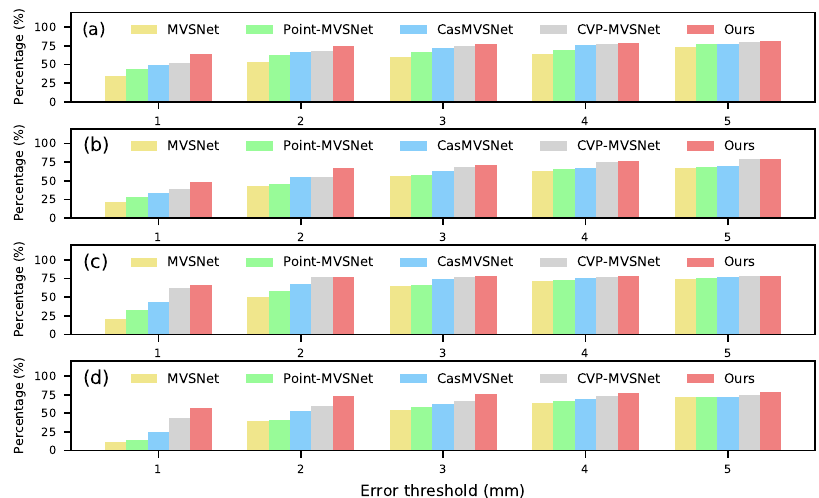}
   \end{overpic}
   \caption{Quantitative comparisons on the depth map prediction of the whole \emph{DTU} test set (a) and the challenging test subsets: \emph{Specular reflection} (b),  \emph{Shadow} (c) and \emph{Occlusion} (d).
   Ours represents the proposed RayMVSNet.
   The percentage (Y-axis) represents the ratio of the pixels whose depth prediction error is smaller than the specific error thresholds (X-axis).}
   \label{fig:depth_curve}
\end{figure}


\begin{table*}
\footnotesize\centering
\caption{
Quantitative results on the \emph{Tanks \& temples} dataset. We use the F-score as the evaluation metric (higher is better).
}
\label{tab:tat}
\begin{tabular}{p{3cm}<{\centering}p{1.05cm}<{\centering}p{1.05cm}<{\centering}p{1.05cm}<{\centering}p{1.5cm}<{\centering}p{1.05cm}<{\centering}p{1.05cm}<{\centering}p{1.05cm}<{\centering}p{1.05cm}<{\centering}p{1.05cm}<{\centering}}
\toprule
Method & Family & Francis & Horse & Light house & M60 & Panther & Playground & Train & Mean \\ \midrule
MVSNet~\cite{yao2018mvsnet} & 55.99 & 28.55 & 25.07 & 50.79 & 53.96 & 50.86 & 47.90 & 34.69 & 43.48 \\
R-MVSNet~\cite{yao2019recurrent} & 69.96 & 46.65 & 32.59 & 42.95 & 51.88 & 48.80 & 52.00 & 42.38 & 48.40 \\
PVA-MVSNet~\cite{yi2020pyramid} & 69.36 & 46.80 & 46.01 & 55.74 & 57.23 & 54.75 & 56.70 & 49.06 & 54.46 \\
CVP-MVSNet~\cite{yang2020cost} & 76.50 & 47.74 & 36.34 & 55.12 & 57.28 & 54.28 & 57.43 & 47.54 & 54.03 \\
CasMVSNet~\cite{gu2020cascade} & 76.37 & 58.45 & 46.26 & 55.81 & 56.11 & 54.06 & 58.18 & 49.51 & 56.84 \\
UCS-Net~\cite{cheng2020deep} & 76.09 & 53.16 &	43.03 & 54.00 & 55.60 & 51.49 & 57.38 & 47.89 & 54.83 \\
D2HC-RMVSNet~\cite{yan2020dense} & 74.69 & 56.04 & \textbf{49.42} & \textbf{60.08} & 59.81 & 59.61 & \textbf{60.04} & \textbf{53.92} & 59.20 \\
U-MVS~\cite{xu2021digging} & 76.49 & 60.04 & 49.20 & 55.52 & 55.33 & 51.22 & 56.77 & 52.63 & 57.15 \\ \midrule
RayMVSNet & \textbf{78.55} & \textbf{61.93} & 45.48 & 57.59 & \textbf{61.00} & \textbf{59.78} & 59.19 & 52.32 & \textbf{59.48}\\
RayMVSNet++ & 77.82 & 60.10 & 44.51 & \underline{58.21} & 58.32 & 57.23 & \underline{59.20} & \underline{52.34} & 58.47\\
\bottomrule
\end{tabular}
\end{table*}

\begin{table*}
\footnotesize\centering
\caption{
Quantitative results on the \emph{ScanNet} dataset. We use multiple metrics to comprehensively evaluate our method and several baselines on depth estimation.
}
\label{tab:scannet_depth}
\begin{tabular}{p{1.9cm}<{\centering}p{1.4cm}<{\centering}p{1.4cm}<{\centering}p{1.4cm}<{\centering}p{1.4cm}<{\centering}p{1.4cm}<{\centering}p{1.6cm}<{\centering}p{1.6cm}<{\centering}p{1.6cm}<{\centering}}
\toprule
Method & AbsRel$\downarrow$&	SqRel$\downarrow$&	Log10$\downarrow$&	RMSE$\downarrow$&	RMSELog$\downarrow$&   $\delta<{1.25}\uparrow$& $\delta<{1.25}^{2}\uparrow$&$\delta<{1.25}^{3}\uparrow$ \\ \midrule
Bts~\cite{lee2019big}& 0.117&	0.052&	0.049&	0.270&	0.151&	0.862&	0.966&	0.992   \\
Bts$^{*}$~\cite{lee2019big}& 0.088&	0.035&	0.038&	0.228&	0.128&	0.916&	0.980&	0.994   \\
MVSNet~\cite{yao2018mvsnet}& 0.098&	0.046&	0.042&	0.256&	0.143&	0.893&	0.968&	0.989   \\
CasMVSNet~\cite{gu2020cascade}& 0.088&	0.039&	0.039&	0.251&	0.128&	0.912&	0.976&	0.992   \\
UCS-Net~\cite{cheng2020deep}& 0.075&	0.029&	0.032&	0.197&	0.107&	0.936&	0.984&	0.995   \\
MVS2D~\cite{yang2022mvs2d}& 0.069&	0.022&	0.030&	0.188&	0.097&	0.951&	0.990&	\textbf{0.998}   \\\midrule
RayMVSNet & 0.074 & 0.029 & 0.037 & 0.195 & 0.107 & 0.947 & 0.986 & 0.996 \\
RayMVSNet++& \textbf{0.058}&	\textbf{0.016}&	\textbf{0.025}&	\textbf{0.157}&	\textbf{0.085}&	\textbf{0.963}&	\textbf{0.993}&	\textbf{0.998}   \\
\bottomrule
\end{tabular}
\end{table*}

\begin{table}[t]
\footnotesize\centering
\caption{
Evaluation metrics for depth estimation. $d_i$ and $d_i^*$ denote the estimated depth and ground truth, respectively. $N$ is the pixel number of the depth image.
}
\label{tab:depth_metric}
\setlength{\tabcolsep}{4pt}
\begin{tabular}{p{2.5cm}<{\centering}p{4.5cm}<{\centering}}
\hline
\toprule
\textbf{Metric} & \textbf{Formulation} \\ \midrule
AbsRel & $\frac{1}{N}\sum_{i}\frac{|d_{i}-d_{i}^{*}|}{d_{i}^{*}} $ \\
SqRel &$\frac{1}{N}\sum_{i}\frac{(d_{i}-d_{i}^{*})^{2}}{d_{i}^{*}} $ \\
Log10 &$\frac{1}{N}\sum_{i}|{{\log_{10}{d_{i}}-\log_{10}{d_{i}^{*}}}}| $  \\
RMSE& $\sqrt{\frac{1}{N}\sum_{i}({{d_{i}-d_{i}^{*}}})^{2}} $   \\
RMSELog& $\sqrt{\frac{1}{N}\sum_{i}({{\log{d_{i}}-\log{d_{i}^{*}}}})^{2}} $  \\
$\delta<{1.25}^{k}$& $\frac{1}{N}\sum_{i}(\max(\frac{d_{i}}{d_{i}^{*}},\frac{d_{i}^{*}}{d_{i}})<{1.25}^{k})$ \\
Percentage@x& $\frac{1}{N}\sum_{i}I(|d_{i}-d_{i}^{*}|<x)$  \\
\hline
\end{tabular}
\end{table}


\begin{figure*}[t] \centering
	\begin{overpic}[width=1.0\linewidth,tics=10]{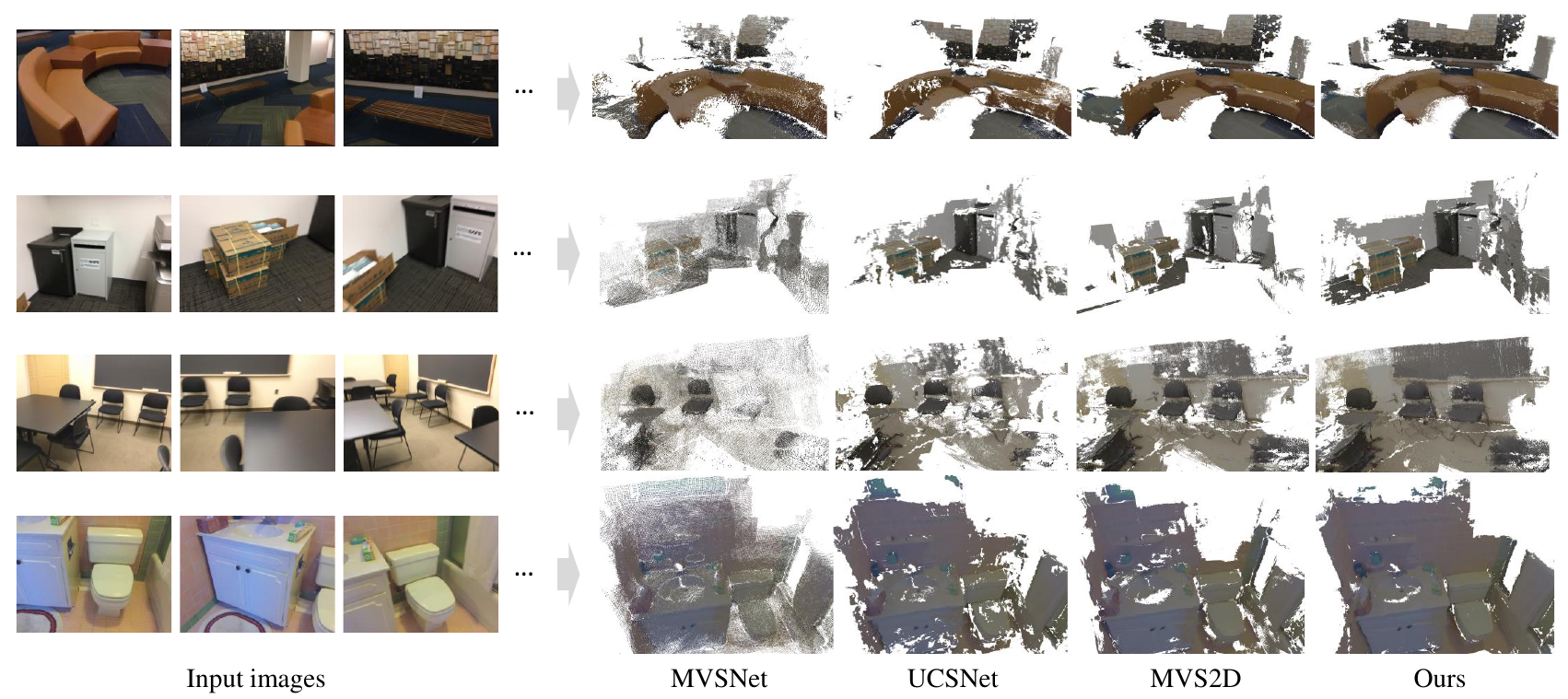}
   \end{overpic}
   \caption{Visual comparisons on the reconstructed point cloud on the \emph{ScanNet} dataset~\cite{dai2017scannet}. RayMVSNet++ achieves better reconstruction results in terms of both accuracy and completeness, thanks to the local-frustum-based context aggregation which introduces more context to tolerate the imperfectness of the input images.}
   \label{fig:scannet_pc}
\end{figure*} 

\begin{figure}[t] \centering
	\begin{overpic}[width=1.0\linewidth,tics=10]{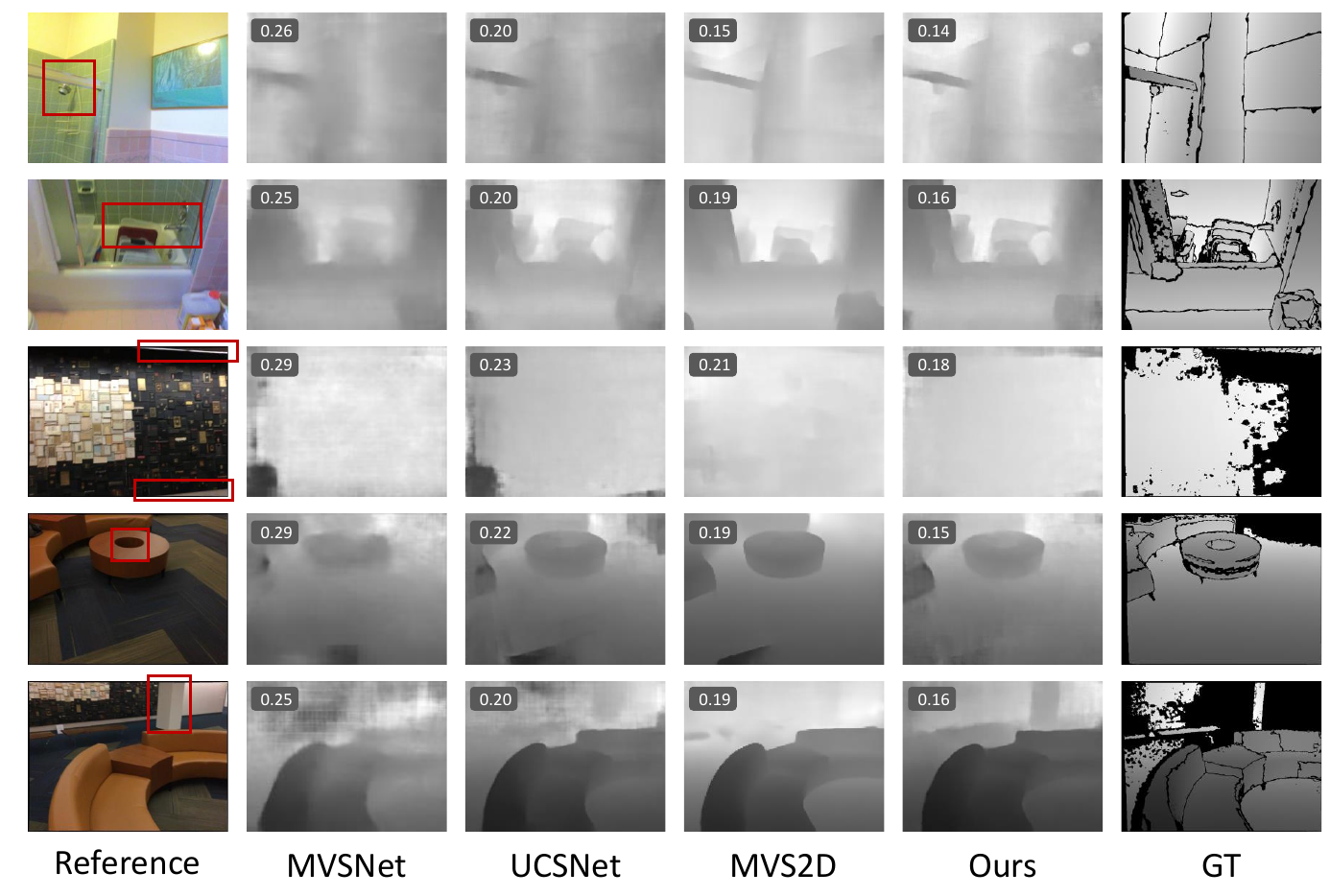}
   \end{overpic}
   \caption{Visual comparisons on the estimated depth on the ScanNet dataset. It shows that RayMVSNet++ achieves more accurate depth estimation in the challenging regions as highlighted. The RMSE (m) is reported on the upper-left of each example.}
   \label{fig:scannet_depth}
\end{figure} 

\begin{figure}[t] \centering
	\begin{overpic}[width=1.0\linewidth,tics=10]{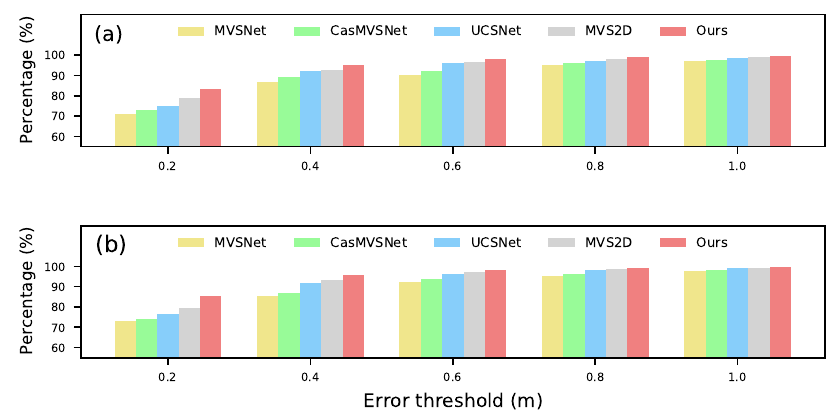}
   \end{overpic}
   \caption{Quantitative comparisons on the depth map prediction of
the challenging test subsets in \emph{ScanNet}: \emph{Textureless} (a), \emph{Large depth variation} (b). Ours represents the proposed RayMVSNet++. The percentage (Y-axis) represents the ratio of the pixels whose depth prediction error is smaller than the specific error thresholds (X-axis).}
   \label{fig:scannet_depth_curve}
\end{figure} 

\begin{figure}[t] \centering
	\begin{overpic}[width=1.0\linewidth,tics=10]{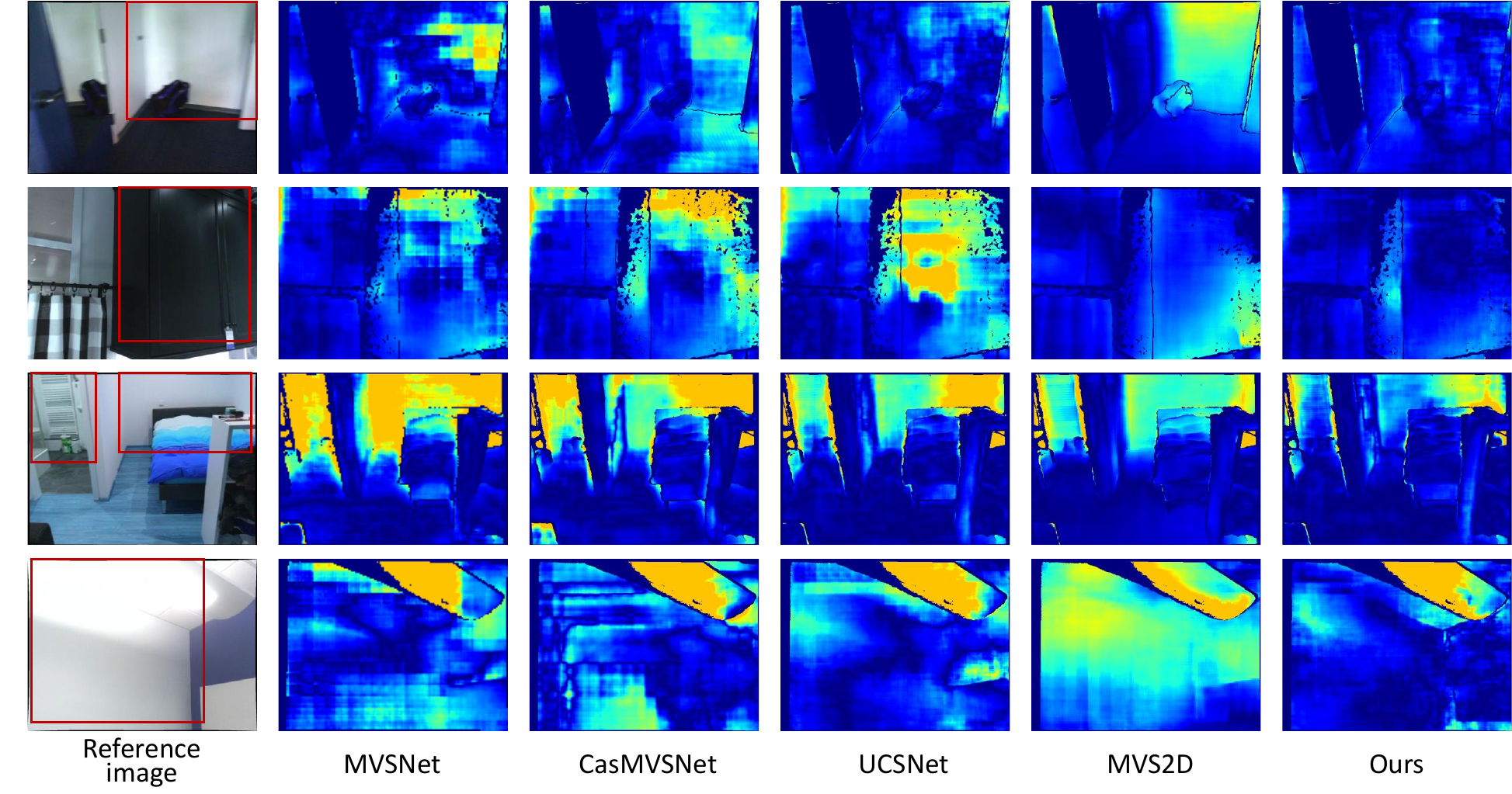}
   \end{overpic}
   \caption{Visual comparisons on the \emph{Textureless} test set in the \emph{ScanNet} dataset.}
   \label{fig:scannet_subset1}
\end{figure}


\begin{figure}[t] \centering
	\begin{overpic}[width=1.0\linewidth,tics=10]{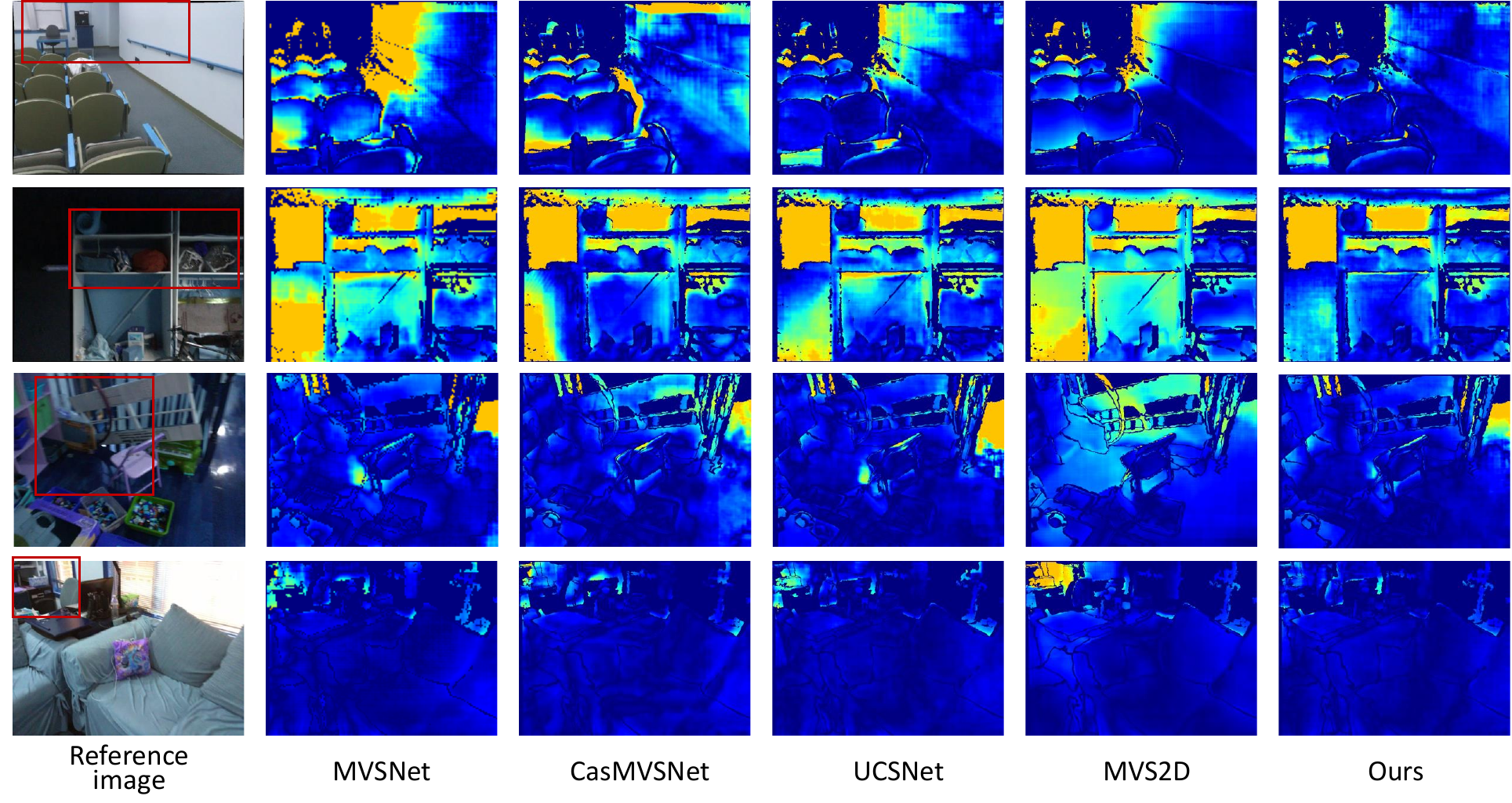}
   \end{overpic}
   \caption{Visual comparisons on the \emph{Large depth variation} test set in the \emph{ScanNet} dataset.}
   \label{fig:scannet_subset2}
\end{figure}


\noindent\textbf{Evaluation on reconstructed point cloud.}\
To evaluate RayMVSNet on \emph{DTU}.
we compare \emph{Accuracy \& Completeness} of the reconstructed point cloud.
The quantitative results are shown in Table \ref{tab:dtu}. It shows that our method not only produces competitive results in terms of \emph{Accuracy} and \emph{Completeness}, but also achieves the state-of-the-art \emph{Overall} performance.
This demonstrates the effectiveness of our method, especially on balancing the trade-off between \emph{Accuracy} and \emph{Completeness}.
The qualitative comparisons are visualized in Figure \ref{fig:visual_comp}. It is shown that our method achieves high-quality reconstruction in various scenarios. In particular, our method outperforms the baselines in scenes with textureless regions, heavy occlusion, and complex geometry.

\noindent\textbf{Evaluation on challenging regions.}\
To further demonstrate our advantage, we compare RayMVSNet with existing works, in terms of the predicted depth map.
The quantitative comparisons on the whole \emph{DTU} test set (Figure \ref{fig:depth_curve} a) and the challenging subsets (Figure \ref{fig:depth_curve} b-d) are reported.
The Percentage$@$x metric is used.
The percentage (Y-axis) represents the ratio of the pixels whose depth prediction error is smaller than the specific error thresholds (X-axis). Higher percentages represent better performances.
It is clear that our method outperforms all the baselines in all error thresholds.
Crucially, our method is more general and robust in challenging cases as shown in Figure \ref{fig:depth}, thanks to the prior learnt from the ray-based 1D implicit field.

\subsection{Performance on Tanks \& Temples}
We compare our method with the baselines on \emph{Tanks \& Temples}. Following the protocol of previous work~\cite{gu2020cascade}, we use the network trained on \emph{DTU}. \emph{F-score} is the evaluation metric. The quantitative results are shown in Table \ref{tab:tat}.
RayMVSNet achieves the best performance, demonstrating the generality of epipolar transformer and ray-based 1D implicit field on large-scale scenes.
\revise{RayMVSNet++ outperforms RayMVSNet on several test scenes while maintaining comparable mean performance to RayMVSNet. This is because most of the images in Tanks \& Temples are captured under good conditions, e.g. in sufficient and stable lighting conditions without motion blur, which RayMVSNet is sufficient to handle.}
\revise{RayMVSNet++ is inferior to the baseline of D2HC-RMVSNet~\cite{yan2020dense} in the scenes with large planar regions, such as Horse and Light house. The reason might be that D2HC-RMVSNet adopted a hybrid recurrent regularization module on the cost volume which provides a mechanism to implicitly involve the structural prior of planar regions for performance improvements.}

\subsection{Performance on ScanNet}

\noindent\textbf{Evaluation on depth estimation.}\
We first evaluate our method in terms of depth estimation on \emph{ScanNet}. The dataset contains low-quality images, so it is especially suitable for the evaluation of the proposed local-frustum-based context aggregation.
The evaluation metrics for depth estimation were adopted.
In this experiment, we set the receptive field $t$ of the local-frustum-based context aggregation as $9$ for RayMVSNet++.
The results are reported in Table~\ref{tab:scannet_depth}.
We see RayMVSNet++ achieves the best performance in all metrics over existing methods.
This demonstrates that RayMVSNet++ could tolerate the imperfection on the input images, i.e. motion blur or inferior lighting conditions.
\revise{In particular, RayMVSNet++ outperforms RayMVSNet, confirming our motivation of developing RayMVSNet++, i.e. aggregating context on challenging regions and low-quality images exist.}
The visual comparisons on depth estimation are provided in Figure~\ref{fig:scannet_depth}.

\begin{table}[t]
\footnotesize\centering
\caption{
\revise{Quantitative results on the \emph{ScanNet} dataset. We evaluate the reconstructed point cloud using the distance metric~\cite{aanaes2016large}. The numbers are reported in $m$ (lower is better).
}}
\label{tab:scannet_pointcloud}
\begin{tabular}{cccc}
\toprule
Method & Accuracy & Completeness & Overall \\ \midrule
Bts$^{*}$~\cite{lee2019big} & 0.048 & 0.213 & 0.130\\
MVSNet~\cite{yao2018mvsnet} & 0.037 & 0.187 & 0.112\\
UCS-Net~\cite{cheng2020deep} & 0.034 & 0.142 & 0.088\\
MVS2D~\cite{yang2022mvs2d} & \textbf{0.031} & 0.147 & 0.089\\ \midrule
RayMVSNet & 0.034 & 0.153 & 0.094\\
RayMVSNet++ & 0.032 & \textbf{0.138} & \textbf{0.083}\\ \bottomrule
\end{tabular}
\end{table}

\revise{
\noindent\textbf{Evaluation on reconstructed point cloud.}\
We also evaluate the quality of the reconstructed point cloud produced by our method.
The \emph{Accuracy \& Completeness} metrics are adopted.
As reported in Table~\ref{tab:scannet_pointcloud}, RayMVSNet++ achieves the best overall performance, which is consistent with the evaluation on depth estimation.
We provide examples of visual comparisons on the reconstructed point cloud in Figure~\ref{fig:scannet_pc}.
}

\noindent\textbf{Evaluation on challenging regions.}\
We also study how our method performs in challenging regions to understand its effectiveness better.
The experiment is conducted on the two subsets, i.e. \emph{Textureless} and \emph{Large depth variation}.
Percentage$@$x is the evaluation metric.
Figure~\ref{fig:scannet_depth_curve} reports the results.
We see RayMVSNet++ outperforms all the baselines in all test subsets with all error thresholds (X-axis).
Notably, RayMVSNet++ outperforms the state-of-the-art methods by a large margin with an error threshold $0.2mm$.
We visualize the examples of the challenging regions in Figure~\ref{fig:scannet_subset1} and~\ref{fig:scannet_subset2}. Note that the depth estimation on the highlighted regions is extremely difficult due to the textureless surfaces and the large depth variation.

\subsection{Ablation Study}
In Table \ref{tab:ablation}, we conduct ablation studies to quantify the efficacy of several crucial components in RayMVSNet and RayMVSNet++.
Unless specifically mentioned otherwise, the experiments are conducted on the \emph{DTU} dataset.

\noindent\textbf{Feature aggregation.}\
The cross-view feature aggregation is a key component of RayMVSNet. To evaluate the importance, we compare the full method to several baselines without some specific component: \emph{w/o epipolar transformer}, \emph{w/o 2D image feature} and \emph{w/o 3D volume feature}.
\revise{To be specific, \emph{w/o epipolar transformer} denotes the baseline that discards the epipolar transformer and uses the fetched multi-view features $\bF^I_{p}$ instead of the aggregated attention-aware feature of epipolar transformer $\bF^A_{p}$ in equation~\ref{eq:point_feature2}. \emph{w/o 2D image features} represents the baseline that discards the multi-view 2D image feature $\bF^A_{\mu,p}$, $\bF^A_{\sigma,p}$, and $\bF^A_{1,p}$ in equation~\ref{eq:point_feature2}. \emph{w/o 3D features} is the baseline that discards the 3D volume feature $\bF_p^V$ in equation~\ref{eq:point_feature2}.}
It clearly shows that all these baselines make the performance decline.
It is worth noting that \emph{w/o epipolar transformer} achieves a lower completeness score, indicating epipolar transformer could make the reconstruction complete by providing more reliable cross-view correlations.
We also compare our epipolar transformer to other multi-view feature aggregation methods. In the experiment of \emph{vis-max feature aggregation}, we replace the epipolar transformer with the visibility-aware max-pooling feature aggregation~\cite{chen2020visibility}. The result indicates epipolar transformer is a better solution.

\revise{
\noindent\textbf{Ray-based inference.}\
Our method learns the 1D implicit field by the ray-based inference. To show its necessity, a straightforward baseline is to learn the implicit field in the 3D space of the reference frustum, such that there is no ray-based inference. This baseline adopts the same cross-view feature aggregation as the full method, and predicts the SDF of sampled points in the reference frustum by using an MLP. The depth map is then generated by a ray-casting algorithm from the predicted SDFs.
}
Unsurprisingly, experiments show this network is hard to converge and leads to low quantitative performance, which suggests that the ray-based 1D implicit field indeed simplifies the learning and is suitable to the MVS problem.

\begin{table}
\footnotesize\centering
\caption{
Ablation studies of RayMVSNet. The performance under distance metric is reported (lower is better).
}
\label{tab:ablation}
\begin{tabular}{cccc}
\toprule
Method  & Accuracy & Completeness & Overall \\ \midrule
w/o epipolar transformer & 0.347 & 0.339 & 0.343 \\
w/o 2D image feature & 0.345 & 0.352 & 0.348 \\
w/o 3D volume feature & 0.434 & 0.322 & 0.378 \\
vis-max feature aggregation & 0.345 & 0.331 & 0.338 \\
w/o ray-based inference & 0.573 & 0.642 & 0.608 \\
Ray with Transformer & \textbf{0.339} & 0.343 & 0.341  \\
Ray with average pooling & 0.356 & 0.406 & 0.381 \\
Ray with max pooling & 0.466 & 0.383 & 0.424 \\
w/o SDF prediction & 0.354 & 0.330 & 0.342 \\
Visibility-aware view aggregation & 0.345 & 0.331 & 0.338 \\ \midrule
RayMVSNet & 0.341 & \textbf{0.319} & \textbf{0.330} \\
\bottomrule
\end{tabular}
\end{table}

\begin{table}
\footnotesize\centering
\caption{
Ablation studies of RayMVSNet++. p@x represents Percentage@x.
}
\label{tab:ablation2}
\begin{tabular}{ccccc}
\toprule
{Method} & RMSE(m)$\downarrow$& p@0.2$\uparrow$& p@0.4$\uparrow$&p@$0.6\uparrow$ \\ \midrule
{w/o frustum}& 0.211&	0.794&	0.918&	0.963  \\
{w/o gating unit}& 0.193&	0.807&	0.925&	0.966   \\
{w/o Gumbel-Softmax}& 0.176&	0.838&	0.950&	0.980   \\ \midrule
{RayMVSNet++}& \textbf{0.158}&	\textbf{0.861}&	\textbf{0.957}&	\textbf{0.982}   \\
\bottomrule
\end{tabular}
\end{table}

\noindent\textbf{Other ray-based implicit field models.}\
In order to reveal the need of the proposed LSTM, we compare our method against several baselines with alternative models of processing sequential data. To be specific, we study the effects of replacing the LSTM with average pooling, max pooling, and Transformer~\cite{vaswani2017attention}, respectively. The \emph{Ray with average pooling} and the \emph{Ray with max pooling} baselines aggregate ray feature by average pooling and max pooling over all sampled points, respectively. The aggregated features are then used to predict the zero-crossing location. The point-wise SDF predictions are also performed as an auxiliary task.
The result shows that our method outperforms all the baselines.
In particular, the performance drops significantly with the \emph{Ray with average pooling} and the \emph{Ray with max pooling}, implying that the modeling of ray-based 1D implicit field is a non-trivial task.
The \emph{Ray with Transformer} is inferior to the full method, in terms of the \emph{Overall score}, confirming that LSTM is more appropriate to our problem.

\begin{figure}[t] \centering
	\begin{overpic}[width=1.0\linewidth,tics=10]{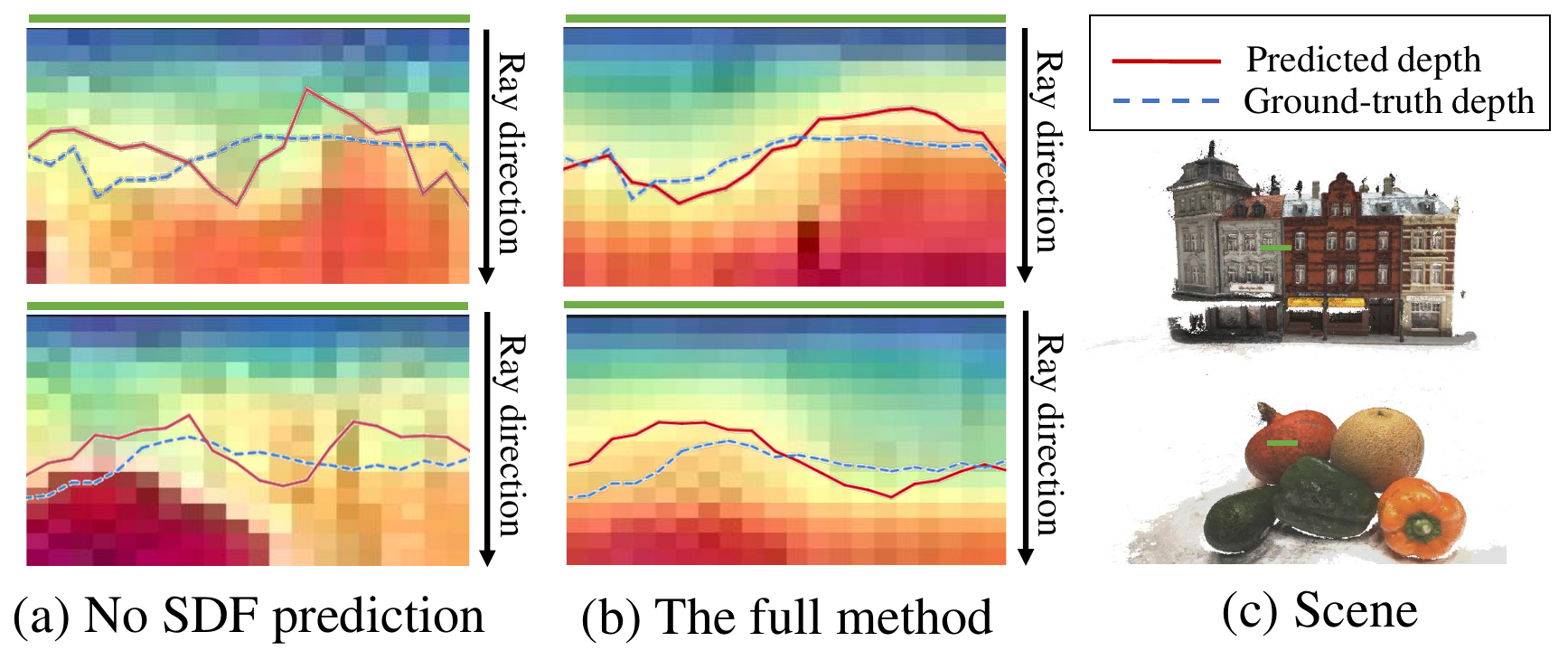}
   \end{overpic}
   \caption{Mid-layer feature map t-SNE visualization of the w/o SDF prediction baseline (a) and the full method (b) for the green segment marked in the scenes in (c). }
   \label{fig:sdf}
\end{figure}

\noindent\textbf{No SDF prediction.}\
The SDF prediction is an auxiliary task in RayMVSNet. We demonstrate its influence by turning it off and comparing to the full method. The performance of \emph{w/o SDF prediction} baseline is inferior to the full method, demonstrating the joint training of SDF prediction and zero-crossing position prediction is indeed helpful, due to the extra supervision of SDF. Examples are visualized in Figure \ref{fig:sdf} which compares the mid-layer features of the full model and the baseline without SDF prediction. We can see that the mid-layer features of the full method, with SDF supervision, maintain a better monotonicity along the ray direction, resulting in more accurate predictions.

\noindent\textbf{Alternative multi-view aggregation.}\
We conduct an experiment on replacing our epipolar transformer with the visibility-aware multi-view feature aggregation method~\cite{chen2020visibility}. The results show that our method outperforms the alternative. This reveals the fact that attention mechanisms are indeed helpful to our multi-view feature aggregation task.


\noindent\textbf{Local-frustum-based context aggregation.}\
Frustum context aggregation is at the core of the proposed method. To reveal the effectiveness, we conduct several ablation studies by removing either the entire module or some key components.
The experiments are conducted on the \emph{ScanNet} dataset.
The results are reported in Table~\ref{tab:ablation2}.
We can make the following conclusions. First, the decline in performance on the baseline without the entire local-frustum-based context aggregation module (\emph{w/o frustum}) indicates the proposed module is necessary. Second, the baseline that uses all context in the receptive field without a selection by the gating unit (\emph{w/o gating unit}) leads to a relatively inferior performance, demonstrating the adaptive context selection is useful. Third, by using the soft decisions during both the forward and backward pass (\emph{w/o Gumbel-Softmax}), the performance drops especially on the $RMSE$ and $p@0.2$ metrics. This is consistent with the conclusions of some recent methods that also unitized the Gumbel-Softmax trick~\cite{veit2017convolutional,kong2019pixel,verelst2020dynamic}.

\subsection{Sensitivity to coarse depth quality}
\begin{figure}[t] \centering
	\begin{overpic}[width=1.0\linewidth,tics=10]{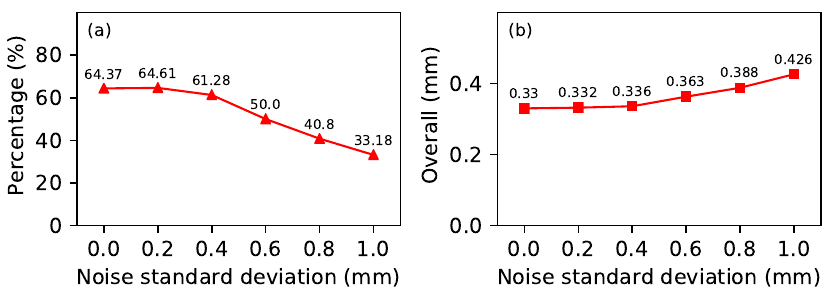}
   \end{overpic}
   \caption{Sensitivity to coarse depth quality. The percentage of pixel-wise depth predictions whose error is smaller than $1mm$ (a) and the overall score of point cloud reconstruction (b) are reported. }
   \label{fig:noise}
\end{figure}

We show our method is robust to the incorrectness of coarse depth prediction by conducting a pressure test. In the experiment, we add Gaussian noise to the predicted coarse depth maps, during both the training and testing phases. We report the performance of the depth map prediction and the point cloud reconstruction on \emph{DTU}. Figure~\ref{fig:noise} shows RayMVSNet is robust to moderate perturbation (noise standard deviation $\leq0.4mm$). It is interesting to see that the quality of depth map prediction slightly increases when moderate noise is added. This demonstrates that data augmentation such as modest perturbation to coarse depth is helpful for training a more generalizable RayMVSNet.
Moreover, we conduct experiments of replacing the MVSNet with other MVSNet variants, e.g., UCS-MVSNet, Fast-MVSNet, and CVP-MVSNet, for coarse depth estimation. We found consistent improvement of depth estimation for the alternative backbones. In particular, our method with a UCS-MVSNet backbone achieves a $0.326$ overall score on the DTU dataset, which is slightly better compared to the original RayMVSNet.


\begin{figure}[t] \centering
	\begin{overpic}[width=1.0\linewidth,tics=10]{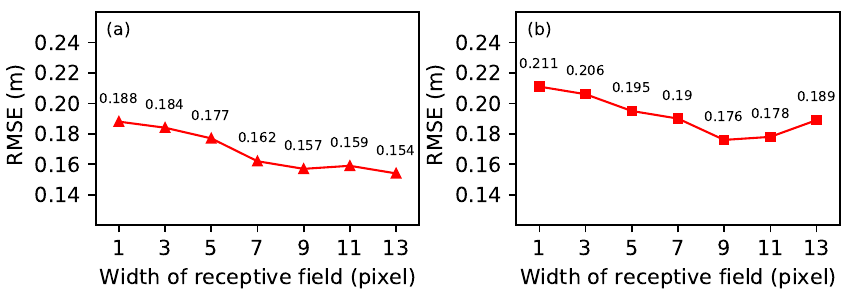}
   \end{overpic}
   \caption{Sensitivity to the width of the receptive field. The RMSE on the \emph{Textureless} and \emph{Large depth variation} test sets are reported.
   In general, we found our method is generally robust and not very sensitive to the width. It achieves the top performance when the width of the receptive field $t=13$ and $9$, respectively.
   }
   \label{fig:sensitivity_t}
\end{figure}

\subsection{Sensitivity to width of receptive field}
\begin{figure}[t] \centering
	\begin{overpic}[width=1.0\linewidth,tics=10]{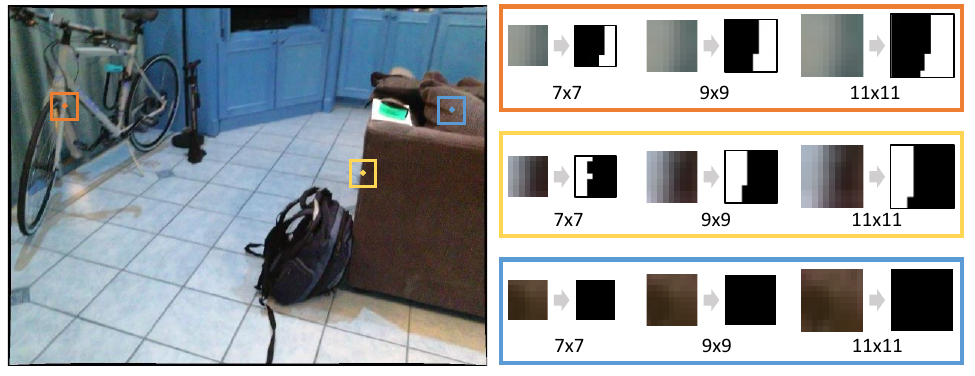}
   \end{overpic}
   \caption{The effectiveness of the gating unit in the local-frustum-based context aggregation. We see the gating unit is able to successfully select the semantically relevant pixels with different width of the receptive field.}
   \label{fig:frustum_vis}
\end{figure}

We also test RayMVSNet++ using different widths of the receptive field in the local-frustum-based context aggregation. We train our method on \emph{ScanNet} with different width and test the trained models on the \emph{Textureless} and \emph{Large depth variation} test sets. The RMSE are showed in Figure~\ref{fig:sensitivity_t}.
When $t=1$, the method essentially equals to the original RayMVSNet~\cite{xi2022raymvsnet}.
It shows that the local-frustum-based context aggregation is indeed helpful on \emph{ScanNet} with more challenging examples.
In particular, RayMVSNet++ is robust when $t\geq7$, demonstrating our method is not sensitive to the parameter. It achieves the top performance when $t = 13$ and $9$, respectively. It shows that the context across large neighboring pixels is more significant to the depth estimation in the textureless regions.
Figure~\ref{fig:frustum_vis} provides some examples of how the attentional gating unit performs with different widths of the receptive field.
We have also tried increasing the width of the receptive field on \emph{DTU}. However, we do not see significant performance improvements. This verifies the idea that the local-frustum-based context aggregation is only helpful to challenging datasets with low-quality images caused by poor lighting conditions or motion blur.


\begin{figure}[t] \centering
	\begin{overpic}[width=1.0\linewidth,tics=10]{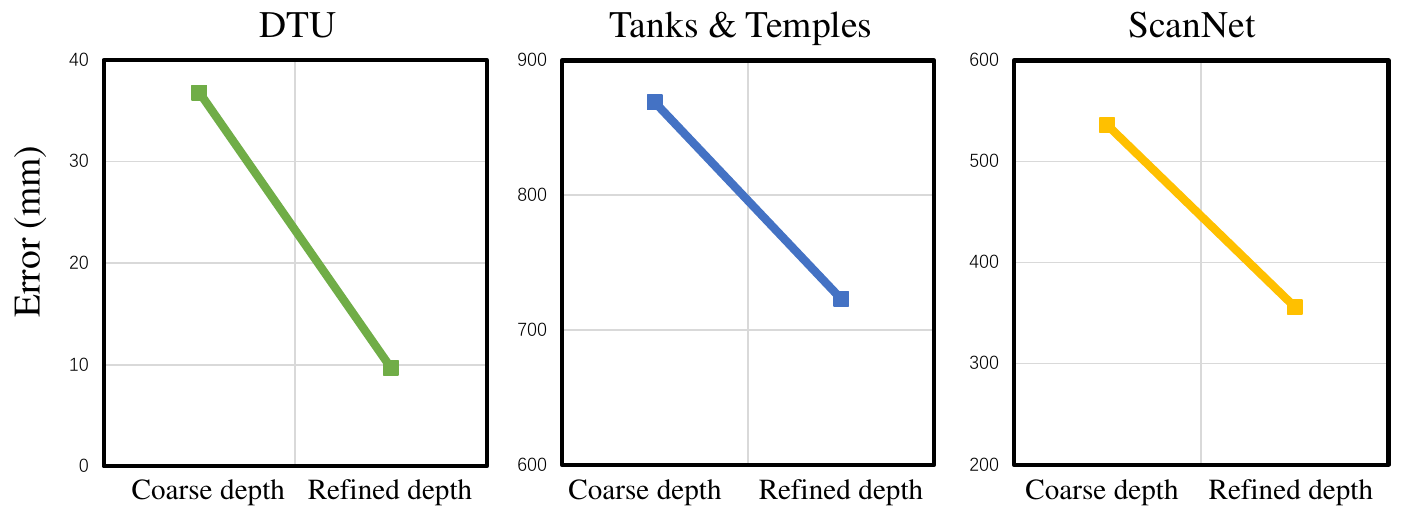}
   \end{overpic}
   \caption{\revise{The errors in depth estimation on the rays whose ground-truth depth is outside the enlarged searching region. We see that ray-based inference is able to improve the accuracy, demonstrating its ability to handle inaccurate coarse depth estimation.
   }}
   \label{fig:outlier}
\end{figure}

\revise{
\subsection{Handling inaccurate coarse depth}
Despite the conservative parameter settings, a small proportion of the true depth might fall outside of the search space induced by the estimated coarse depth.
Although such cases are the minority ($<3\%$), our method is able to alleviate this problem by estimating the relative location $l$ on the ray.
In such cases, during the ray-based inference, the estimated relative location $l$ would be outside the enlarged searching region $[0,1]$.
Since those cases exist in both the training and testing phases, our method is able to learn to estimate those by the regression in equation~\ref{eq:ratio}.
Figure~\ref{fig:outlier} provides visualizations of the depth estimation accuracy before and after the ray-based inference on the rays whose ground-truth depth is outside the enlarged searching region.
We see that our method is able to improve the accuracy, demonstrating its ability to handle inaccurate coarse depth estimation.
Note that the errors shown in the figure are determined by both the accuracy of depth estimation itself and the range of the enlarged searching region.
We set the range of the enlarged searching region as $20mm$ in DTU, $1000mm$ in Tanks \& Temples, and $600mm$ in ScanNet.
}

\subsection{Qualitative results}
\begin{figure*}[!t] \centering \begin{overpic}[width=1.0\linewidth,tics=10]{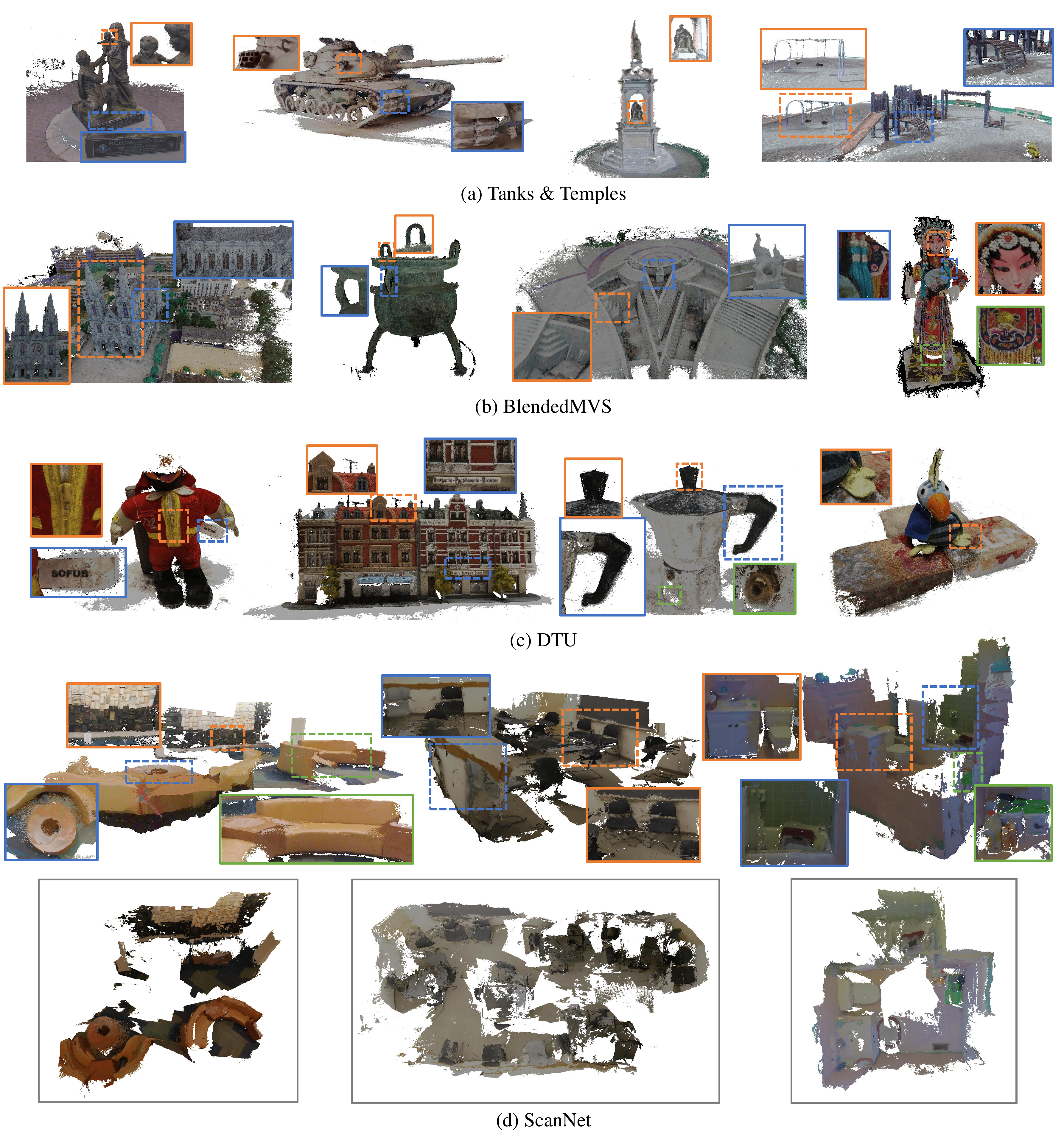}
   \end{overpic}
   \caption{
   \revise{Gallery of the reconstructed point cloud on (a) \emph{Tanks \& temples}, (b) \emph{BlendedMVS}, (c) \emph{DTU}, and (d) \emph{ScanNet}. In (d), the scenes in the rectangles with solid line represents the reconstructed point cloud of the whole scene observed from the top view.}
   }
   \label{fig:gallery}
\end{figure*}

We visualize the qualitative results of our method on several datasets in Figure \ref{fig:gallery}. Note that our method is able to reconstruct large-scale scenes with fine-grained geometry details, such as the highlighted regions.




\section{Conclusion and Discussion}
\label{sec:conclusion}
We have presented RayMVSNet++, which learns to directly optimize the depth value along each camera ray. An epipolar transformer is designed to enable sequential modeling of 1D ray-based implicit fields, which essentially mimics the epipolar line search in traditional MVS. The ray-based approach demonstrates significant performance boost with only a low-res cost volume.
In particular, a local-frustum-based context aggregation is proposed to extend the receptive field of the ray-based model, leading to more accurate and robust predictions.
The method has been demonstrated to be effective on three public datasets, achieving state-of-the-art performance.

Our method has the following limitations. First, although we have demonstrated the method is robust to the coarse depth quality, there is still a small proportion of challenging regions whose depth cannot be accurately estimated due to the large error in the coarse depth prediction.
Second, our method relies on accurate camera poses. For scenarios that do not meet this requirement, our method cannot produce accurate outputs, since it cannot optimize the camera pose and the 3D points simultaneously.

An interesting future direction is to further enhance the ray-based deep MVS approach so that cost volume convolution could be completely saved.
In most deep MVS works, 3D point cloud is recovered from the estimated depth map as post-processing. Therefore, we would also like to study the end-to-end optimization of 3D point clouds~\cite{shi2020symmetrynet}.
\revise{Moreover, our method assumes the camera poses are given, it is interesting to explore estimating the camera pose~\cite{cai2021pose} and reconstructing scene/object surfaces in a uniform framework, such that the two tasks would boost each other.} 

\appendices
\setcounter{equation}{0}

\ifCLASSOPTIONcompsoc
  \section*{Acknowledgments}
\else
  \section*{Acknowledgment}
\fi

We thank the anonymous reviewers for their valuable comments.
This work was supported in part by the National Key Research and Development Program of China under Grants 2018AAA0102200 and 2018YFB1305100, in part by NSFC under Grants 62132021 and 62002379, in part by Natural Science Foundation of Hunan Province of China under Grants 2023JJ20051 and 2023JJ20048.

\ifCLASSOPTIONcaptionsoff
  \newpage
\fi

\bibliographystyle{ieee_fullname}
\bibliography{egbib}

\begin{thebibliography}{10}\itemsep=-1pt

\bibitem{aanaes2016large}
Henrik Aan{\ae}s, Rasmus~Ramsb{\o}l Jensen, George Vogiatzis, Engin Tola, and
  Anders~Bjorholm Dahl.
\newblock Large-scale data for multiple-view stereopsis.
\newblock {\em International Journal of Computer Vision}, 120(2):153--168,
  2016.

\bibitem{andrew2001multiple}
Alex~M Andrew.
\newblock Multiple view geometry in computer vision.
\newblock {\em Kybernetes}, 2001.

\bibitem{cai2021pose}
Qi Cai, Lilian Zhang, Yuanxin Wu, Wenxian Yu, and Dewen Hu.
\newblock A pose-only solution to visual reconstruction and navigation.
\newblock {\em IEEE Transactions on Pattern Analysis and Machine Intelligence},
  45(1):73--86, 2023.

\bibitem{carion2020end}
Nicolas Carion, Francisco Massa, Gabriel Synnaeve, Nicolas Usunier, Alexander
  Kirillov, and Sergey Zagoruyko.
\newblock End-to-end object detection with transformers.
\newblock In {\em European conference on computer vision}, pages 213--229.
  Springer, 2020.

\bibitem{chabra2020deep}
Rohan Chabra, Jan~E Lenssen, Eddy Ilg, Tanner Schmidt, Julian Straub, Steven
  Lovegrove, and Richard Newcombe.
\newblock Deep local shapes: Learning local sdf priors for detailed 3d
  reconstruction.
\newblock In {\em European Conference on Computer Vision}, pages 608--625.
  Springer, 2020.

\bibitem{chen2021mvsnerf}
Anpei Chen, Zexiang Xu, Fuqiang Zhao, Xiaoshuai Zhang, Fanbo Xiang, Jingyi Yu,
  and Hao Su.
\newblock Mvsnerf: Fast generalizable radiance field reconstruction from
  multi-view stereo.
\newblock In {\em Proceedings of the IEEE/CVF International Conference on
  Computer Vision}, pages 14124--14133, 2021.

\bibitem{chen2019point}
Rui Chen, Songfang Han, Jing Xu, and Hao Su.
\newblock Point-based multi-view stereo network.
\newblock In {\em Proceedings of the IEEE/CVF International Conference on
  Computer Vision}, pages 1538--1547, 2019.

\bibitem{chen2020visibility}
Rui Chen, Songfang Han, Jing Xu, and Hao Su.
\newblock Visibility-aware point-based multi-view stereo network.
\newblock {\em IEEE transactions on pattern analysis and machine intelligence},
  43(10):3695--3708, 2020.

\bibitem{chen2019learning}
Zhiqin Chen and Hao Zhang.
\newblock Learning implicit fields for generative shape modeling.
\newblock In {\em Proceedings of the IEEE/CVF Conference on Computer Vision and
  Pattern Recognition}, pages 5939--5948, 2019.

\bibitem{cheng2020deep}
Shuo Cheng, Zexiang Xu, Shilin Zhu, Zhuwen Li, Li~Erran Li, Ravi Ramamoorthi,
  and Hao Su.
\newblock Deep stereo using adaptive thin volume representation with
  uncertainty awareness.
\newblock In {\em Proceedings of the IEEE/CVF Conference on Computer Vision and
  Pattern Recognition}, pages 2524--2534, 2020.

\bibitem{dai2017scannet}
Angela Dai, Angel~X Chang, Manolis Savva, Maciej Halber, Thomas Funkhouser, and
  Matthias Nie{\ss}ner.
\newblock Scannet: Richly-annotated 3d reconstructions of indoor scenes.
\newblock In {\em Proceedings of the IEEE conference on computer vision and
  pattern recognition}, pages 5828--5839, 2017.

\bibitem{dai2019second}
Tao Dai, Jianrui Cai, Yongbing Zhang, Shu-Tao Xia, and Lei Zhang.
\newblock Second-order attention network for single image super-resolution.
\newblock In {\em Proceedings of the IEEE/CVF conference on computer vision and
  pattern recognition}, pages 11065--11074, 2019.

\bibitem{deng2021deformed}
Yu Deng, Jiaolong Yang, and Xin Tong.
\newblock Deformed implicit field: Modeling 3d shapes with learned dense
  correspondence.
\newblock In {\em Proceedings of the IEEE/CVF Conference on Computer Vision and
  Pattern Recognition}, pages 10286--10296, 2021.

\bibitem{ding2022transmvsnet}
Yikang Ding, Wentao Yuan, Qingtian Zhu, Haotian Zhang, Xiangyue Liu, Yuanjiang
  Wang, and Xiao Liu.
\newblock Transmvsnet: Global context-aware multi-view stereo network with
  transformers.
\newblock In {\em Proceedings of the IEEE/CVF Conference on Computer Vision and
  Pattern Recognition}, pages 8585--8594, 2022.

\bibitem{dosovitskiy2020image}
Alexey Dosovitskiy, Lucas Beyer, Alexander Kolesnikov, Dirk Weissenborn,
  Xiaohua Zhai, Thomas Unterthiner, Mostafa Dehghani, Matthias Minderer, Georg
  Heigold, and Sylvain~and Gelly.
\newblock An image is worth 16x16 words: Transformers for image recognition at
  scale.
\newblock In {\em International Conference on Learning Representations}, 2021.

\bibitem{duan2020curriculum}
Yueqi Duan, Haidong Zhu, He Wang, Li Yi, Ram Nevatia, and Leonidas~J Guibas.
\newblock Curriculum deepsdf.
\newblock In {\em European Conference on Computer Vision}, pages 51--67.
  Springer, 2020.

\bibitem{galliani2015massively}
Silvano Galliani, Katrin Lasinger, and Konrad Schindler.
\newblock Massively parallel multiview stereopsis by surface normal diffusion.
\newblock In {\em Proceedings of the IEEE International Conference on Computer
  Vision}, pages 873--881, 2015.

\bibitem{genova2020local}
Kyle Genova, Forrester Cole, Avneesh Sud, Aaron Sarna, and Thomas Funkhouser.
\newblock Local deep implicit functions for 3d shape.
\newblock In {\em Proceedings of the IEEE/CVF Conference on Computer Vision and
  Pattern Recognition}, pages 4857--4866, 2020.

\bibitem{gu2020cascade}
Xiaodong Gu, Zhiwen Fan, Siyu Zhu, Zuozhuo Dai, Feitong Tan, and Ping Tan.
\newblock Cascade cost volume for high-resolution multi-view stereo and stereo
  matching.
\newblock In {\em Proceedings of the IEEE/CVF Conference on Computer Vision and
  Pattern Recognition}, pages 2495--2504, 2020.

\bibitem{hartmann2017learned}
Wilfried Hartmann, Silvano Galliani, Michal Havlena, Luc Van~Gool, and Konrad
  Schindler.
\newblock Learned multi-patch similarity.
\newblock In {\em Proceedings of the IEEE International Conference on Computer
  Vision}, pages 1586--1594, 2017.

\bibitem{he2020epipolar}
Yihui He, Rui Yan, Katerina Fragkiadaki, and Shoou-I Yu.
\newblock Epipolar transformer for multi-view human pose estimation.
\newblock In {\em Proceedings of the IEEE/CVF Conference on Computer Vision and
  Pattern Recognition Workshops}, pages 1036--1037, 2020.

\bibitem{hochreiter1997long}
Sepp Hochreiter and J{\"u}rgen Schmidhuber.
\newblock Long short-term memory.
\newblock {\em Neural computation}, 9(8):1735--1780, 1997.

\bibitem{huang2018deepmvs}
Po-Han Huang, Kevin Matzen, Johannes Kopf, Narendra Ahuja, and Jia-Bin Huang.
\newblock Deepmvs: Learning multi-view stereopsis.
\newblock In {\em Proceedings of the IEEE Conference on Computer Vision and
  Pattern Recognition}, pages 2821--2830, 2018.

\bibitem{im2019dpsnet}
Sunghoon Im, Hae-Gon Jeon, Stephen Lin, and In~So Kweon.
\newblock Dpsnet: End-to-end deep plane sweep stereo.
\newblock In {\em International Conference on Learning Representations}, 2019.

\bibitem{jang2016categorical}
Eric Jang, Shixiang Gu, and Ben Poole.
\newblock Categorical reparameterization with gumbel-softmax.
\newblock In {\em International Conference on Learning Representations}, 2017.

\bibitem{ji2017surfacenet}
Mengqi Ji, Juergen Gall, Haitian Zheng, Yebin Liu, and Lu Fang.
\newblock Surfacenet: An end-to-end 3d neural network for multiview stereopsis.
\newblock In {\em Proceedings of the IEEE International Conference on Computer
  Vision}, pages 2307--2315, 2017.

\bibitem{jiang2020local}
Chiyu Jiang, Avneesh Sud, Ameesh Makadia, Jingwei Huang, Matthias Nie{\ss}ner,
  Thomas Funkhouser, et~al.
\newblock Local implicit grid representations for 3d scenes.
\newblock In {\em Proceedings of the IEEE/CVF Conference on Computer Vision and
  Pattern Recognition}, pages 6001--6010, 2020.

\bibitem{kar2017learning}
Abhishek Kar, Christian H\"ane, and Jitendra Malik.
\newblock Learning a multi-view stereo machine.
\newblock In {\em Advances in neural information processing systems}. 2017.

\bibitem{katharopoulos2020transformers}
Angelos Katharopoulos, Apoorv Vyas, Nikolaos Pappas, and Fran{\c{c}}ois
  Fleuret.
\newblock Transformers are rnns: Fast autoregressive transformers with linear
  attention.
\newblock In {\em International Conference on Machine Learning}, pages
  5156--5165. PMLR, 2020.

\bibitem{kitaev2020reformer}
Nikita Kitaev, {\L}ukasz Kaiser, and Anselm Levskaya.
\newblock Reformer: The efficient transformer.
\newblock In {\em International Conference on Learning Representations}, 2020.

\bibitem{knapitsch2017tanks}
Arno Knapitsch, Jaesik Park, Qian-Yi Zhou, and Vladlen Koltun.
\newblock Tanks and temples: Benchmarking large-scale scene reconstruction.
\newblock {\em ACM Transactions on Graphics (ToG)}, 36(4):1--13, 2017.

\bibitem{kong2019pixel}
Shu Kong and Charless Fowlkes.
\newblock Pixel-wise attentional gating for scene parsing.
\newblock In {\em 2019 IEEE Winter Conference on Applications of Computer
  Vision (WACV)}, pages 1024--1033. IEEE, 2019.

\bibitem{lee2019set}
Juho Lee, Yoonho Lee, Jungtaek Kim, Adam Kosiorek, Seungjin Choi, and Yee~Whye
  Teh.
\newblock Set transformer: A framework for attention-based
  permutation-invariant neural networks.
\newblock In {\em International conference on machine learning}, pages
  3744--3753. PMLR, 2019.

\bibitem{lee2019big}
Jin~Han Lee, Myung-Kyu Han, Dong~Wook Ko, and Il~Hong Suh.
\newblock From big to small: Multi-scale local planar guidance for monocular
  depth estimation.
\newblock {\em arXiv preprint arXiv:1907.10326}, 2019.

\bibitem{li2019attention}
Yanwei Li, Xinze Chen, Zheng Zhu, Lingxi Xie, Guan Huang, Dalong Du, and
  Xingang Wang.
\newblock Attention-guided unified network for panoptic segmentation.
\newblock In {\em Proceedings of the IEEE/CVF Conference on Computer Vision and
  Pattern Recognition}, pages 7026--7035, 2019.

\bibitem{liu2022neural}
Yuan Liu, Sida Peng, Lingjie Liu, Qianqian Wang, Peng Wang, Christian Theobalt,
  Xiaowei Zhou, and Wenping Wang.
\newblock Neural rays for occlusion-aware image-based rendering.
\newblock In {\em Proceedings of the IEEE/CVF Conference on Computer Vision and
  Pattern Recognition}, pages 7824--7833, 2022.

\bibitem{luo2019p}
Keyang Luo, Tao Guan, Lili Ju, Haipeng Huang, and Yawei Luo.
\newblock P-mvsnet: Learning patch-wise matching confidence aggregation for
  multi-view stereo.
\newblock In {\em Proceedings of the IEEE/CVF International Conference on
  Computer Vision}, pages 10452--10461, 2019.

\bibitem{luo2020attention}
Keyang Luo, Tao Guan, Lili Ju, Yuesong Wang, Zhuo Chen, and Yawei Luo.
\newblock Attention-aware multi-view stereo.
\newblock In {\em Proceedings of the IEEE/CVF Conference on Computer Vision and
  Pattern Recognition}, pages 1590--1599, 2020.

\bibitem{ma2021epp}
Xinjun Ma, Yue Gong, Qirui Wang, Jingwei Huang, Lei Chen, and Fan Yu.
\newblock Epp-mvsnet: Epipolar-assembling based depth prediction for multi-view
  stereo.
\newblock In {\em Proceedings of the IEEE/CVF International Conference on
  Computer Vision}, pages 5732--5740, 2021.

\bibitem{mescheder2019occupancy}
Lars Mescheder, Michael Oechsle, Michael Niemeyer, Sebastian Nowozin, and
  Andreas Geiger.
\newblock Occupancy networks: Learning 3d reconstruction in function space.
\newblock In {\em Proceedings of the IEEE/CVF Conference on Computer Vision and
  Pattern Recognition}, pages 4460--4470, 2019.

\bibitem{mildenhall2020nerf}
Ben Mildenhall, Pratul~P Srinivasan, Matthew Tancik, Jonathan~T Barron, Ravi
  Ramamoorthi, and Ren Ng.
\newblock Nerf: Representing scenes as neural radiance fields for view
  synthesis.
\newblock In {\em European conference on computer vision}, pages 405--421.
  Springer, 2020.

\bibitem{murez2020atlas}
Zak Murez, Tarrence Van~As, James Bartolozzi, Ayan Sinha, Vijay Badrinarayanan,
  and Andrew Rabinovich.
\newblock Atlas: End-to-end 3d scene reconstruction from posed images.
\newblock In {\em Computer Vision--ECCV 2020: 16th European Conference,
  Glasgow, UK, August 23--28, 2020, Proceedings, Part VII 16}, pages 414--431.
  Springer, 2020.

\bibitem{nelson2020learning}
Henry~J Nelson and Nikolaos Papanikolopoulos.
\newblock Learning continuous object representations from point cloud data.
\newblock In {\em 2020 IEEE/RSJ International Conference on Intelligent Robots
  and Systems (IROS)}, pages 2446--2451. IEEE, 2020.

\bibitem{niemeyer2020differentiable}
Michael Niemeyer, Lars Mescheder, Michael Oechsle, and Andreas Geiger.
\newblock Differentiable volumetric rendering: Learning implicit 3d
  representations without 3d supervision.
\newblock In {\em Proceedings of the IEEE/CVF Conference on Computer Vision and
  Pattern Recognition}, pages 3504--3515, 2020.

\bibitem{park2019deepsdf}
Jeong~Joon Park, Peter Florence, Julian Straub, Richard Newcombe, and Steven
  Lovegrove.
\newblock Deepsdf: Learning continuous signed distance functions for shape
  representation.
\newblock In {\em Proceedings of the IEEE/CVF Conference on Computer Vision and
  Pattern Recognition}, pages 165--174, 2019.

\bibitem{peng2020convolutional}
Songyou Peng, Michael Niemeyer, Lars Mescheder, Marc Pollefeys, and Andreas
  Geiger.
\newblock Convolutional occupancy networks.
\newblock In {\em Computer Vision--ECCV 2020: 16th European Conference,
  Glasgow, UK, August 23--28, 2020, Proceedings, Part III 16}, pages 523--540.
  Springer, 2020.

\bibitem{qin2022geometric}
Zheng Qin, Hao Yu, Changjian Wang, Yulan Guo, Yuxing Peng, and Kai Xu.
\newblock Geometric transformer for fast and robust point cloud registration.
\newblock In {\em Proceedings of the IEEE/CVF Conference on Computer Vision and
  Pattern Recognition}, pages 11143--11152, 2022.

\bibitem{saito2019pifu}
Shunsuke Saito, Zeng Huang, Ryota Natsume, Shigeo Morishima, Angjoo Kanazawa,
  and Hao Li.
\newblock Pifu: Pixel-aligned implicit function for high-resolution clothed
  human digitization.
\newblock In {\em Proceedings of the IEEE/CVF international conference on
  computer vision}, pages 2304--2314, 2019.

\bibitem{seitz2006comparison}
Steven~M Seitz, Brian Curless, James Diebel, Daniel Scharstein, and Richard
  Szeliski.
\newblock A comparison and evaluation of multi-view stereo reconstruction
  algorithms.
\newblock In {\em 2006 IEEE computer society conference on computer vision and
  pattern recognition (CVPR'06)}, volume~1, pages 519--528. IEEE, 2006.

\bibitem{shaw2018self}
Peter Shaw, Jakob Uszkoreit, and Ashish Vaswani.
\newblock Self-attention with relative position representations.
\newblock {\em arXiv preprint arXiv:1803.02155}, 2018.

\bibitem{shi2020symmetrynet}
Yifei Shi, Junwen Huang, Hongjia Zhang, Xin Xu, Szymon Rusinkiewicz, and Kai
  Xu.
\newblock Symmetrynet: learning to predict reflectional and rotational
  symmetries of 3d shapes from single-view rgb-d images.
\newblock {\em ACM Transactions on Graphics (TOG)}, 39(6):1--14, 2020.

\bibitem{silberman2012indoor}
Nathan Silberman, Derek Hoiem, Pushmeet Kohli, and Rob Fergus.
\newblock Indoor segmentation and support inference from rgbd images.
\newblock In {\em European conference on computer vision}, pages 746--760.
  Springer, 2012.

\bibitem{sitzmann2020metasdf}
Vincent Sitzmann, Eric~R Chan, Richard Tucker, Noah Snavely, and Gordon
  Wetzstein.
\newblock Metasdf: Meta-learning signed distance functions.
\newblock In {\em Advances in neural information processing systems}, 2020.

\bibitem{sun2021loftr}
Jiaming Sun, Zehong Shen, Yuang Wang, Hujun Bao, and Xiaowei Zhou.
\newblock Loftr: Detector-free local feature matching with transformers.
\newblock In {\em Proceedings of the IEEE/CVF conference on computer vision and
  pattern recognition}, pages 8922--8931, 2021.

\bibitem{sun2021neuralrecon}
Jiaming Sun, Yiming Xie, Linghao Chen, Xiaowei Zhou, and Hujun Bao.
\newblock Neuralrecon: Real-time coherent 3d reconstruction from monocular
  video.
\newblock In {\em Proceedings of the IEEE/CVF Conference on Computer Vision and
  Pattern Recognition}, pages 15598--15607, 2021.

\bibitem{takikawa2021neural}
Towaki Takikawa, Joey Litalien, Kangxue Yin, Karsten Kreis, Charles Loop, Derek
  Nowrouzezahrai, Alec Jacobson, Morgan McGuire, and Sanja Fidler.
\newblock Neural geometric level of detail: Real-time rendering with implicit
  3d shapes.
\newblock In {\em Proceedings of the IEEE/CVF Conference on Computer Vision and
  Pattern Recognition}, pages 11358--11367, 2021.

\bibitem{tretschk2020patchnets}
Edgar Tretschk, Ayush Tewari, Vladislav Golyanik, Michael Zollh{\"o}fer,
  Carsten Stoll, and Christian Theobalt.
\newblock Patchnets: Patch-based generalizable deep implicit 3d shape
  representations.
\newblock In {\em European Conference on Computer Vision}, pages 293--309.
  Springer, 2020.

\bibitem{vaswani2017attention}
Ashish Vaswani, Noam Shazeer, Niki Parmar, Jakob Uszkoreit, Llion Jones,
  Aidan~N Gomez, {\L}ukasz Kaiser, and Illia Polosukhin.
\newblock Attention is all you need.
\newblock In {\em Advances in neural information processing systems}, pages
  5998--6008, 2017.

\bibitem{veit2017convolutional}
Andreas Veit and Serge Belongie.
\newblock Convolutional networks with adaptive inference graphs.
\newblock In {\em European Conference on Computer Vision}, 2018.

\bibitem{verelst2020dynamic}
Thomas Verelst and Tinne Tuytelaars.
\newblock Dynamic convolutions: Exploiting spatial sparsity for faster
  inference.
\newblock In {\em Proceedings of the IEEE/CVF Conference on Computer Vision and
  Pattern Recognition}, pages 2320--2329, 2020.

\bibitem{wang2022generalizable}
Dan Wang, Xinrui Cui, Septimiu Salcudean, and Z~Jane Wang.
\newblock Generalizable neural radiance fields for novel view synthesis with
  transformer.
\newblock {\em arXiv preprint arXiv:2206.05375}, 2022.

\bibitem{wang2021patchmatchnet}
Fangjinhua Wang, Silvano Galliani, Christoph Vogel, Pablo Speciale, and Marc
  Pollefeys.
\newblock Patchmatchnet: Learned multi-view patchmatch stereo.
\newblock In {\em Proceedings of the IEEE/CVF Conference on Computer Vision and
  Pattern Recognition}, pages 14194--14203, 2021.

\bibitem{wang2021neus}
Peng Wang, Lingjie Liu, Yuan Liu, Christian Theobalt, Taku Komura, and Wenping
  Wang.
\newblock Neus: Learning neural implicit surfaces by volume rendering for
  multi-view reconstruction.
\newblock {\em Advances in Neural Information Processing Systems}, 2021.

\bibitem{wang2021ibrnet}
Qianqian Wang, Zhicheng Wang, Kyle Genova, Pratul~P Srinivasan, Howard Zhou,
  Jonathan~T Barron, Ricardo Martin-Brualla, Noah Snavely, and Thomas
  Funkhouser.
\newblock Ibrnet: Learning multi-view image-based rendering.
\newblock In {\em Proceedings of the IEEE/CVF Conference on Computer Vision and
  Pattern Recognition}, pages 4690--4699, 2021.

\bibitem{wei2021nerfingmvs}
Yi Wei, Shaohui Liu, Yongming Rao, Wang Zhao, Jiwen Lu, and Jie Zhou.
\newblock Nerfingmvs: Guided optimization of neural radiance fields for indoor
  multi-view stereo.
\newblock In {\em Proceedings of the IEEE/CVF International Conference on
  Computer Vision}, pages 5610--5619, 2021.

\bibitem{wei2021aa}
Zizhuang Wei, Qingtian Zhu, Chen Min, Yisong Chen, and Guoping Wang.
\newblock Aa-rmvsnet: Adaptive aggregation recurrent multi-view stereo network.
\newblock In {\em Proceedings of the IEEE/CVF International Conference on
  Computer Vision}, pages 6187--6196, 2021.

\bibitem{xi2022raymvsnet}
Junhua Xi, Yifei Shi, Yijie Wang, Yulan Guo, and Kai Xu.
\newblock Raymvsnet: Learning ray-based 1d implicit fields for accurate
  multi-view stereo.
\newblock In {\em Proceedings of the IEEE/CVF Conference on Computer Vision and
  Pattern Recognition}, pages 8595--8605, 2022.

\bibitem{xu2021digging}
Hongbin Xu, Zhipeng Zhou, Yali Wang, Wenxiong Kang, Baigui Sun, Hao Li, and Yu
  Qiao.
\newblock Digging into uncertainty in self-supervised multi-view stereo.
\newblock In {\em Proceedings of the IEEE/CVF International Conference on
  Computer Vision}, pages 6078--6087, 2021.

\bibitem{xu2020learning}
Qingshan Xu and Wenbing Tao.
\newblock Learning inverse depth regression for multi-view stereo with
  correlation cost volume.
\newblock In {\em Proceedings of the AAAI Conference on Artificial
  Intelligence}, volume~34, pages 12508--12515, 2020.

\bibitem{xu2019disn}
Qiangeng Xu, Weiyue Wang, Duygu Ceylan, Radomir Mech, and Ulrich Neumann.
\newblock Disn: Deep implicit surface network for high-quality single-view 3d
  reconstruction.
\newblock {\em Advances in neural information processing systems}, 32, 2019.

\bibitem{yan2020dense}
Jianfeng Yan, Zizhuang Wei, Hongwei Yi, Mingyu Ding, Runze Zhang, Yisong Chen,
  Guoping Wang, and Yu-Wing Tai.
\newblock Dense hybrid recurrent multi-view stereo net with dynamic consistency
  checking.
\newblock In {\em European Conference on Computer Vision}, pages 674--689.
  Springer, 2020.

\bibitem{yang2020cost}
Jiayu Yang, Wei Mao, Jose~M Alvarez, and Miaomiao Liu.
\newblock Cost volume pyramid based depth inference for multi-view stereo.
\newblock In {\em Proceedings of the IEEE/CVF Conference on Computer Vision and
  Pattern Recognition}, pages 4877--4886, 2020.

\bibitem{yang2022mvs2d}
Zhenpei Yang, Zhile Ren, Qi Shan, and Qixing Huang.
\newblock Mvs2d: Efficient multi-view stereo via attention-driven 2d
  convolutions.
\newblock In {\em Proceedings of the IEEE/CVF Conference on Computer Vision and
  Pattern Recognition}, pages 8574--8584, 2022.

\bibitem{yao2018mvsnet}
Yao Yao, Zixin Luo, Shiwei Li, Tian Fang, and Long Quan.
\newblock Mvsnet: Depth inference for unstructured multi-view stereo.
\newblock In {\em Proceedings of the European Conference on Computer Vision
  (ECCV)}, pages 767--783, 2018.

\bibitem{yao2019recurrent}
Yao Yao, Zixin Luo, Shiwei Li, Tianwei Shen, Tian Fang, and Long Quan.
\newblock Recurrent mvsnet for high-resolution multi-view stereo depth
  inference.
\newblock In {\em Proceedings of the IEEE/CVF Conference on Computer Vision and
  Pattern Recognition}, pages 5525--5534, 2019.

\bibitem{yariv2021volume}
Lior Yariv, Jiatao Gu, Yoni Kasten, and Yaron Lipman.
\newblock Volume rendering of neural implicit surfaces.
\newblock {\em Advances in Neural Information Processing Systems},
  34:4805--4815, 2021.

\bibitem{yi2020pyramid}
Hongwei Yi, Zizhuang Wei, Mingyu Ding, Runze Zhang, Yisong Chen, Guoping Wang,
  and Yu-Wing Tai.
\newblock Pyramid multi-view stereo net with self-adaptive view aggregation.
\newblock In {\em European Conference on Computer Vision}, pages 766--782.
  Springer, 2020.

\bibitem{yu2021attention}
Anzhu Yu, Wenyue Guo, Bing Liu, Xin Chen, Xin Wang, Xuefeng Cao, and Bingchuan
  Jiang.
\newblock Attention aware cost volume pyramid based multi-view stereo network
  for 3d reconstruction.
\newblock {\em ISPRS Journal of Photogrammetry and Remote Sensing},
  175:448--460, 2021.

\bibitem{yu2020fast}
Zehao Yu and Shenghua Gao.
\newblock Fast-mvsnet: Sparse-to-dense multi-view stereo with learned
  propagation and gauss-newton refinement.
\newblock In {\em Proceedings of the IEEE/CVF Conference on Computer Vision and
  Pattern Recognition}, pages 1949--1958, 2020.

\bibitem{zhang2020visibility}
Jingyang Zhang, Yao Yao, Shiwei Li, Zixin Luo, and Tian Fang.
\newblock Visibility-aware multi-view stereo network.
\newblock {\em British Machine Vision Conference (BMVC)}, 2020.

\bibitem{zhang2021learning}
Jingyang Zhang, Yao Yao, and Long Quan.
\newblock Learning signed distance field for multi-view surface reconstruction.
\newblock In {\em Proceedings of the IEEE/CVF International Conference on
  Computer Vision}, pages 6525--6534, 2021.

\bibitem{zhang2021pa}
Ke Zhang, Mengyu Liu, Jinlai Zhang, and Zhenbiao Dong.
\newblock Pa-mvsnet: Sparse-to-dense multi-view stereo with pyramid attention.
\newblock {\em IEEE Access}, 9:27908--27915, 2021.

\bibitem{zhang2020nerf++}
Kai Zhang, Gernot Riegler, Noah Snavely, and Vladlen Koltun.
\newblock Nerf++: Analyzing and improving neural radiance fields.
\newblock {\em arXiv preprint arXiv:2010.07492}, 2020.

\bibitem{zhang2021long}
Xudong Zhang, Yutao Hu, Haochen Wang, Xianbin Cao, and Baochang Zhang.
\newblock Long-range attention network for multi-view stereo.
\newblock In {\em Proceedings of the IEEE/CVF Winter Conference on Applications
  of Computer Vision}, pages 3782--3791, 2021.

\bibitem{zhao2018triangle}
Yawei Zhao, Kai Xu, En Zhu, Xinwang Liu, Xinzhong Zhu, and Jianping Yin.
\newblock Triangle lasso for simultaneous clustering and optimization in graph
  datasets.
\newblock {\em IEEE Transactions on Knowledge and Data Engineering},
  31(8):1610--1623, 2018.

\bibitem{zheng2021deep}
Zerong Zheng, Tao Yu, Qionghai Dai, and Yebin Liu.
\newblock Deep implicit templates for 3d shape representation.
\newblock In {\em Proceedings of the IEEE/CVF Conference on Computer Vision and
  Pattern Recognition}, pages 1429--1439, 2021.

\bibitem{zhu2021multi}
Jie Zhu, Bo Peng, Wanqing Li, Haifeng Shen, Zhe Zhang, and Jianjun Lei.
\newblock Multi-view stereo with transformer.
\newblock {\em arXiv preprint arXiv:2112.00336}, 2021.

\end{thebibliography}

\begin{IEEEbiography}[{\includegraphics[width=1in,height=1.25in,clip,keepaspectratio]{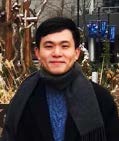}}]{Yifei Shi}
is an Associate Professor at the College of Intelligence Science and Technology, National University of Defense Technology (NUDT). He received his Ph.D. degree in computer science from NUDT in 2019. During 2017-2018, he was a visiting student research collaborator at Princeton University, advised by Thomas Funkhouser and Szymon Rusinkiewicz. His research interests mainly include computer vision, computer graphics, especially on object/scene analysis and manipulation by machine learning and geometric processing techniques. He has published 20+ papers in top-tier conferences and journals, including CVPR, ECCV, ICCV, SIGGRAPH Asia, IEEE Transactions on Pattern Analysis and Machine Intelligence, and ACM Transactions on Graphics.
\end{IEEEbiography}

\begin{IEEEbiography}[{\includegraphics[width=1in,height=1.25in,clip,keepaspectratio]{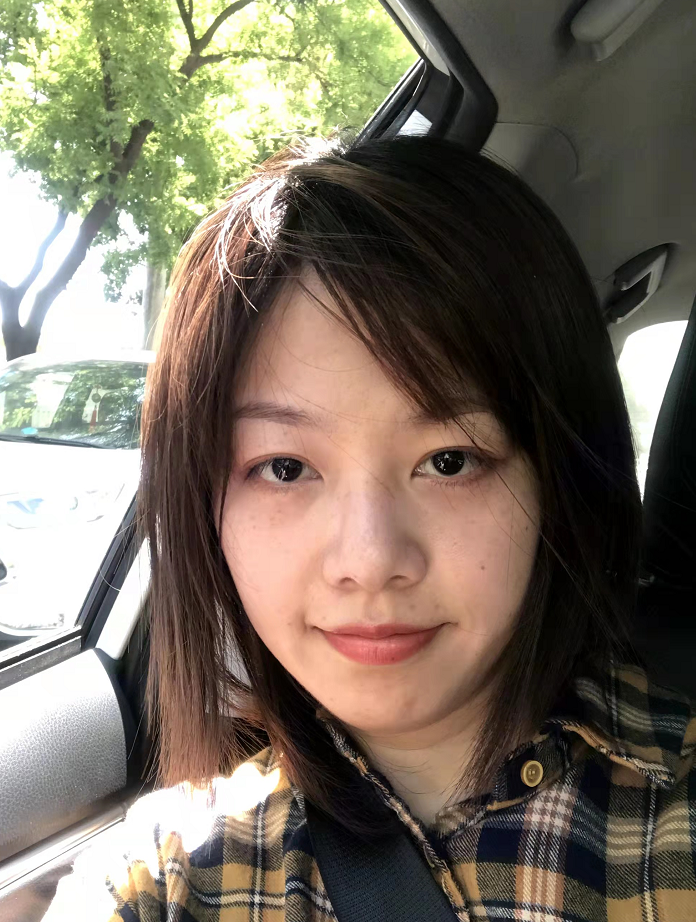}}]{Junhua Xi}
is pursuing her Ph.D. degree in College of Computer Science and Technology, NUDT. She obtained her master's degree from NUDT in 2013. Her research interests include 3D vision and robotics, especially on object analysis, multi-view stereo and large-scale scene reconstruction.
\end{IEEEbiography}

\begin{IEEEbiography}[{\includegraphics[width=1in,height=1.25in,clip,keepaspectratio]{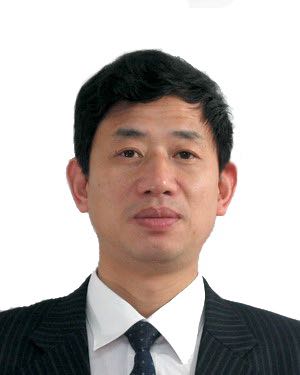}}]{Dewen Hu}
received the B.Sc. and M.Sc. degrees from Xi’an Jiaotong University, China, in 1983 and 1986, respectively, and the Ph.D. degree from the National University of Defense Technology, in 1999. In 1986, he was with the National University of Defense Technology. From October 1995 to October 1996, he was a Visiting Scholar with The University of Sheffield, U.K. In 1996, he was promoted as a Professor. He has authored more than 200 articles in journals, such as the Brain, the Proceedings of the National Academy of Sciences of the United States of America, the NeuroImage, the Human Brain Mapping, the IEEE Transactions on Pattern Analysis and Machine Intelligence, the IEEE Transactions on Image Processing, the IEEE Transactions on Signal Processing, the IEEE Transactions on Neural Networks and Learning Systems, the IEEE Transactions on Medical Imaging, and the IEEE Transactions on Biomedical Engineering. His research interests include pattern recognition and cognitive neuroscience. He is currently an Associate Editor of IEEE Transactions on Systems, Man, and Cybernetics: Systems.
\end{IEEEbiography}

\begin{IEEEbiography}[{\includegraphics[width=1in,height=1.25in,clip,keepaspectratio]{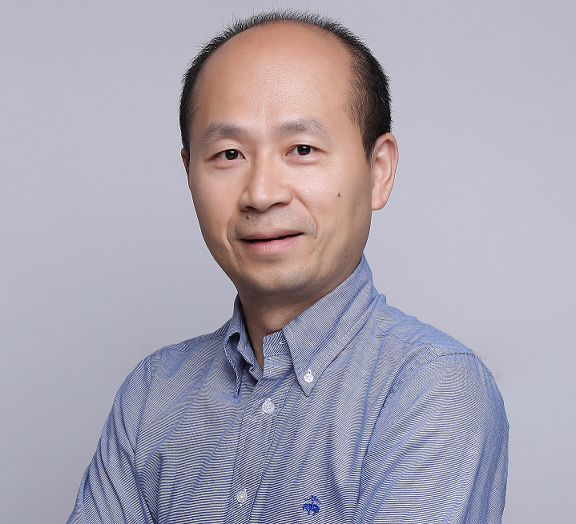}}]{Zhiping Cai}
received the B.Eng., M.A.Sc., and Ph.D. degrees in computer science and technology from the National University of Defense Technology (NUDT), China, in 1996, 2002, and 2005, respectively. He is a full professor in the College of Computer, NUDT. His current research interests include artificial intelligence, network security and big data. He is a senior member of the CCF and a member of the IEEE.
\end{IEEEbiography}

\begin{IEEEbiography}[{\includegraphics[width=1in,height=1.25in,clip,keepaspectratio]{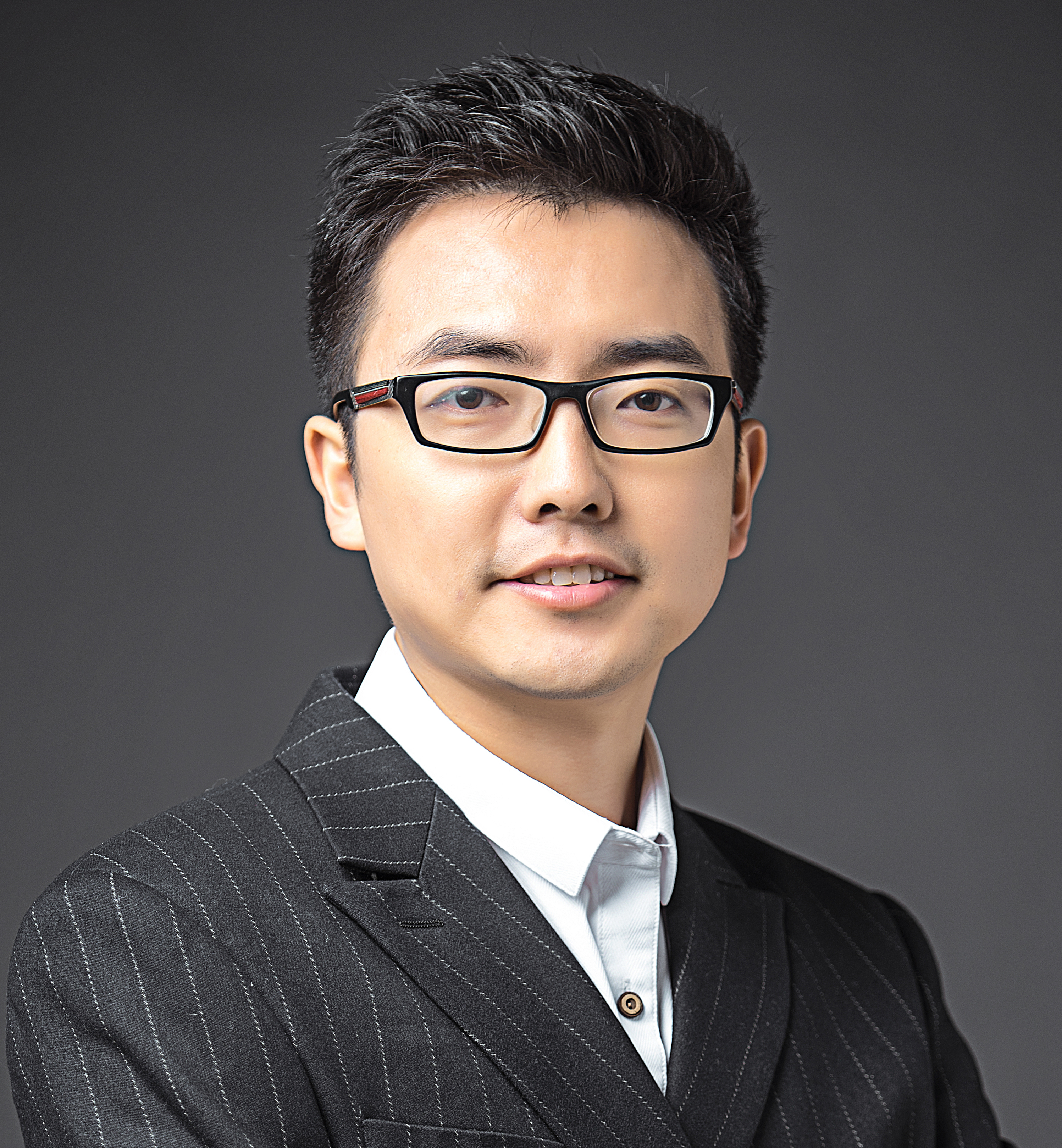}}]{Kai Xu}
is a Professor at the College of Computer, NUDT, where he received his Ph.D. in 2011. He conducted visiting research at Simon Fraser University and Princeton University. His research interests include geometric modeling and shape analysis, especially on data-driven approaches to the problems in those directions, as well as 3D vision and its robotic applications. He has published over 80 research papers, including 20+ SIGGRAPH/TOG papers. He has co-organized several SIGGRAPH Asia courses and Eurographics STAR tutorials. He serves on the editorial board of ACM Transactions on Graphics, Computer Graphics Forum, Computers \& Graphics, and The Visual Computer. He also served as program co-chair of CAD/Graphics 2017, ICVRV 2017 and ISVC 2018, as well as PC member for several prestigious conferences including SIGGRAPH, SIGGRAPH Asia, Eurographics, SGP, PG, etc. His research work can be found in his personal website: www.kevinkaixu.net
\end{IEEEbiography}

\end{document}